\documentclass{article}
\pdfoutput=1
\usepackage{PRIMEarxiv}

\usepackage{amssymb}
\usepackage{pdflscape}
\usepackage{multirow}
\usepackage{booktabs}
\usepackage{caption}
\usepackage[edges]{forest}
\usepackage{tabularx}
\usepackage{graphicx}
\usepackage[round]{natbib}

\usepackage{hyperref}
\usepackage[draft]{todonotes}

\usepackage[framemethod=TikZ]{mdframed}
\usepackage{seqsplit}

\title{Multimodal Document Analytics \\ for Banking Process Automation}

\author{
  \href{https://orcid.org/0000-0002-7579-0751}{\includegraphics[scale=0.06]{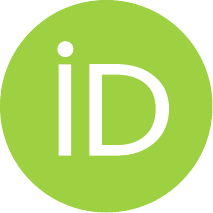}\hspace{1mm}Christopher Gerling} \\
  Chair of Information Systems \\ 
  Humboldt University of Berlin\\  
  Berlin, Germany \\
  \texttt{gerlingc@hu-berlin.de} \\
   \And
 \href{https://orcid.org/0000-0001-7685-262X}{\includegraphics[scale=0.06]{orcid.pdf}\hspace{1mm}Stefan Lessmann} \\
  Chair of Information Systems \\ 
  Humboldt University of Berlin\\  
  Berlin, Germany \\
}

\begin{document}
\maketitle

\begin{abstract}

Traditional banks face increasing competition from FinTechs in the rapidly evolving financial ecosystem. Raising operational efficiency is vital to address this challenge. 
Our study aims to improve the efficiency of document-intensive business processes in banking. To that end, we first review the landscape of business documents in the retail segment. Banking documents often contain text, layout, and visuals, suggesting that document analytics and process automation require more than plain natural language processing (NLP). To verify this and assess the incremental value of visual cues when processing business documents, we compare a recently proposed multimodal model called LayoutXLM to powerful text classifiers (e.g., BERT) and large language models (e.g., GPT) in a case study related to processing company register extracts. \\
The results confirm that incorporating layout information in a model substantially increases its performance.  Interestingly, we also observed that more than 75\% of the best model performance (in terms of the F1 score) can be achieved with as little as 30\% of the training data. This shows that the demand for data labeled data to set up a multi-modal model can be moderate, which simplifies real-world applications of multimodal document analytics. Our study also sheds light on more specific practices in the scope of calibrating a multimodal banking document classifier, including the need for fine-tuning. In sum, the paper contributes original empirical evidence on the effectiveness and efficiency of multi-model models for document processing in the banking business and offers practical guidance on how to unlock this potential in day-to-day operations. 

\end{abstract}


\section{Introduction}
\label{sec:intro}

The banking sector is an integral part of the global economy and plays a critical role in shaping financial growth and stability. In recent years, the banking industry has experienced increasing competition from FinTech companies and the adoption of innovative technologies to drive operational efficiency. Optimizing the wide range of banking processes is an integral part of the digital transformation of banks to remain competitive in a rapidly evolving environment.

Management of the diverse document landscape within core processes, which involve documents of multiple types, structures, and languages, is a key part of banking operations. Advanced document analytics techniques promise to accelerate data extraction, reduce manual processing, and improve the quality and efficiency of decision-making.

The emergence of large foundational language models, such as GPT-4, currently characterizes the field of Natural Language Processing (NLP). This dynamic field currently resembles a ``race of models", with new Deep Neural Network architectures for language processing being published frequently. 
Rather than getting caught up in a battle of benchmarks, this study shifts the focus to real-world applications, specifically, the potential of document analytics to increase efficiency. Some document analytics models provide capabilities beyond text processing. These so-called multimodal models handle documents in which both textual and visual information is crucial for understanding \citep{oral2020}. Examples are scanned invoices \citep{baviskar2021}, money transfer orders forms \citep{oral2020, oral2022} or financial reports with tables and images \citep{kamaruddin2015, pejic2019, dong2023}.

The objectives of this research cover three areas. First, we investigate the diversity and characteristics of the banking document landscape. Next, we intend to develop and validate a proof-of-concept for document analytics, exploring textual and visual feature incorporation.
Lastly, we aim to benchmark our proposed document analytics framework against a ready-to-use large language model, specifically GPT, to elucidate the strengths and limitations of general-purpose models in the complex field of banking document analytics.

Our study offers pivotal contributions to the OR in banking literature. Through our detailed examination of the customer banking document landscape, we reveal unexplored areas for efficiency gains. Further, we demonstrate the importance of including visual features in document analytics models, specifically when using LayoutXLM. Our comparative analysis of LayoutXLM with models such as BERT and GPT provides novel insights into the power of multimodal models in banking contexts.
Addressing real-world obstacles, in particular class imbalance, we uncover a reduced dependency on exhaustive labeled data for deploying pre-trained document analytics models. Finally, we stress the urgency for banking institutions to re-evaluate their analytical paradigms, suggesting a shift towards more advanced frameworks.

\section{Related Work}
\label{sec:related}
This section surveys related studies and sheds light on the state of affairs in two interconnected research areas: NLP applications in banking and the emerging field of multimodal document analytics. 
A list of related studies in both areas including relevant document types in banking operations is summarized in Table \ref{tab:literature-applications}.

\begin{table}[htbp]
  \centering
  \caption{Applications of NLP and Document Analytics in Banking}
  \label{tab:literature-applications}
  \setlength\tabcolsep{5pt}
  \renewcommand{\arraystretch}{1.2}
  \scriptsize
    \begin{tabularx}{\textwidth}{p{90px}Xp{100px}}
    \toprule
    \textbf{Study} & \textbf{Main Topic} & \textbf{Highlighted Documents} \\

    \midrule
    \addlinespace[1ex]
    \multicolumn{3}{l}{\textbf{Natural Language Processing in Banking}} \\
    \addlinespace[1ex]
    \citet{li2011} & Survey of text analysis techniques for corporate disclosures & financial statements, earnings releases, and conference call transcripts   \\
    \citet{Chaturvedi2014} & Sentiment analysis over banking services using online reviews & -\\
    \citet{kamaruddin2015} & Text mining framework for financial statement deviation detection & financial statements  \\
    \citet{kumar2016} & Text mining applications in finance (e.g., FOREX rate prediction) & - \\
    \citet{aureli2017} & Content analysis and text mining for companies' reputation restoration & social and environmental disclosures \\
    \citet{sumathi2017} & Opinion mining using sentiment analysis on social media texts & - \\
    \citet{pejic2019} & Overview of text mining applications in the financial sector & financial statements and legal documents  \\
    \citet{oral2019} & Extract transactions from banking documents using a BiLSTM network & banking orders \\
    \citet{lewis2019} & Highlight the importance of optimizing corporate reports analysis with NLP & balance sheet, income statement, statement of shareholders' equity, and statement of cash flows \\
    \citet{gupta2020} & Review of text mining applications in finance & transaction orders, annual reports, and financial statements \\
    \citet{decaigny2020} & Improving customer churn prediction models using electronic text messages between client and financial advisor & - \\
    \citet{baviskar2021} & Review of AI methods for automated processing of unstructured documents and mention of applications (also in banking) & invoices, passports, ID-cards, diverse application forms, and legal contracts \\
    \citet{sokolov2021} & Transformer-based ESG scoring using social media texts & - \\
    \citet{dong2023} &Keyword extraction to construct an ESG scoring system & financial reports\\
    \addlinespace[1ex]
    \midrule
    \addlinespace[1ex]
    \multicolumn{3}{l}{\textbf{Multimodal Document Analytics in Banking}} \\
    \addlinespace[1ex]
    \citet{engin2019} & Multimodal deep neural networks for classification of Turkish banking documents using textual and visual features & unspecified banking order documents  \\
    \citet{oral2020} & Information extraction from banking documents using deep learning algorithms and word positional features & money transfer orders  \\
    \citet{oral2022} & Impact of different fusion techniques on information extraction from unstructured documents & money transfer orders  \\
    \bottomrule
    \end{tabularx}%
\end{table}

\subsection{NLP Applications in Banking}

In recent years, there has been an increasing number of studies in the banking sector deploying NLP techniques. These studies, which often fall under ``text mining", analyze textual banking data and gain insights through classification, clustering, information extraction, or sentiment analysis \citep{gupta2020}.
Literature reviews \citep{kumar2016, fisher2016, pejic2019, gupta2020, baviskar2021} aggregate different applications of NLP in banking and identify opportunities to use unstructured textual data.

\citet{doumpos2023} categorize AI banking research into some core areas, including bank efficiency and risk management, as well as customer-related insights \citep{doumpos2023}. In the following paragraphs, we dive into these core areas to illustrate NLP applications in banking.

\paragraph{Bank Efficiency} Several studies investigate the role of NLP in improving bank efficiency, defined as achieving the best possible results with the least amount of inputs and costs \citep{doumpos2023}. Bank efficiency can be seen from two perspectives: \textit{cost efficiency}, which seeks to minimize inputs, and \textit{profit efficiency}, which aims to increase outputs \citep{tecles2010}.

The analysis of financial statements and corporate reports, for example, through information extraction or named entity recognition (NER), is a major field where NLP demonstrates efficiency increases. Studies such as \citet{li2011} and \citet{kamaruddin2015} propose the use of NLP to analyze financial statements, helping to optimize processes and reduce costs. 
\citet{lewis2019} underscore NLP's potential to handle massive amounts of data in corporate reporting, uncovering latent attributes and tackling information overload, thus optimizing efficiency. These text-driven approaches can be applied not only to financial statements and corporate reports, but also to various other document types, as discussed in more detail in the section \ref{sec:multimod}.

Customer service also benefits from NLP-enhanced cost efficiency:
Context-aware chatbots and conversational AI provide permanent customer service, handle complex transactions, and adapt to changing customer needs, thus optimizing process efficiency and improving customer experience \citep{suhel2020, mogaji2021, petersson2023}.

On the profit side, NLP enhances revenue streams, particularly in the area of personalized marketing.
Textual data contains insights that can streamline marketing initiatives and personalized product recommendations \citep{pejic2019, chen2020}. For example, \citet{schmitz2023} support stock trading decisions to increase profits using textual data from corporate disclosures.
Furthermore, text analyses on social media can provide more personalized services \citep{sun2018, chen2019}. This is further extended by techniques such as Company2Vec to identify other companies for customer acquisition \citep{gerling2023}. These NLP systems contribute to the offering of customized banking services, leading to business and revenue growth.

\paragraph{Risk Management} Effective risk management is a cornerstone in banking. It requires a complex understanding and accurate quantification of operational, market and credit risk \citep{doumpos2023}. Many studies use NLP techniques to improve risk identification and assessment, using various textual sources such as trade orders, annual reports, and financial statements \citep{gupta2020}. 

For risk assessment and default prediction, text analytics provides innovative methods to improve predictive accuracy. \citet{stevenson2021} develop a combined small business default prediction model using traditional and textual features, highlighting the use of BERT, a transformer-based model known for its success in a variety of NLP tasks, for their textual analysis.  
Similarly, \citet{jiang2018} and \citet{mai2019} improve loan default and bankruptcy prediction by including textual information from descriptive texts or company disclosures.
Furthermore, \citet{netzer2019} find empirical evidence that borrowers who are likely to default use a specific language in their loan application. Including these textual features in a combined model increases default prediction performance.

NLP also seems promising in the area of transaction fraud detection \citep{kotios2022}.
\citet{kotios2022} demonstrate text-based fraud detection, by identifying fraudulent activities in transactions.
In this context, \citet{rodriguez2022} develop a transformer-based model, which outperforms traditional methods in several fraud scenarios.
Furthermore, \citet{craja2020} propose a hierarchical attention network that combines information from financial ratios and managerial comments within corporate annual reports to detect statement fraud.

Financial forecasting is another dimension where NLP is increasingly gaining ground, enhancing the understanding of market risk and volatile underlyings \citep{kumar2016}.
\citet{xing2018} review the emerging field of NLP-based financial forecasting, emphasizing the growing ability of NLP to improve financial market predictions. 
There are advances like Stock2Vec \citep{dang2018}, a two-stream deep learning model for short-term stock forecasting using newspaper articles in combination with historical stock prices. 
Similarly, \citet{li2021} employs an LSTM-based model for stock prediction using online news and fundamental data.

\paragraph{Customer-related Studies} This category includes studies that aim to understand customer behavior without an immediate impact on cost reductions or revenue increases. These additional insights play a critical role in understanding competition, refining product quality, improving customer satisfaction, and thus aligning banking strategies.

In this area, prevention of customer churn and early identification of potential customer loss are of great interest. \citet{decaigny2020} use text messages between financial advisors and clients to enrich customer churn prediction models, achieving the best performance using a combination of textual and structured features within convolutional neural networks. Such models can greatly support customer retention strategies.   

Sentiment analysis and opinion mining also have a considerable impact on banking. \citet{Chaturvedi2014} use automated sentiment analysis to extract customer opinions from bank reviews and to provide insight into strengths and weaknesses of particular areas of a bank (e.g., customer service). Meanwhile, \citet{sumathi2017} use social media text to create a sentiment index for decision support, highlighting the influence of public sentiment on banking strategies.

NLP also assesses companies' environmental and social behavior (ESG) through textual analyses. For example, \citet{aureli2017} applies content analysis to explore firms' reputation restoration strategies after industrial disasters. Then \citet{dong2023} uses keyword extraction from companies' annual financial reports to construct an ESG scoring system. Extending this, \citet{sokolov2021} use BERT to automate ESG scoring from unstructured social media contents, thereby improving ESG risk assessment and creating semi-autonomous ESG rating systems.

\subsection{Multimodal Document Analytics in Banking}
\label{sec:multimod}

NLP research mentions the potential for analyzing banking documents, from traditional financial documents such as financial statements and earnings releases \citep{lewis2019, li2011} to more operationally focused documents such as bank (transaction) orders \citep{oral2019, gupta2020}, ID cards, legal contracts \citep{baviskar2021} or social and environmental disclosures \citep{aureli2017} (more examples in Table \ref{tab:literature-applications}).

Previous research mainly considered documents as ordered collections of word tokens. However, the recent literature perceives banking documents as multidimensional entities that encompass visual elements and layout details \citep{engin2019, oral2020, oral2022, tavakoli2023}.
This perspective acknowledges the variety of visually rich, often multipage, documents that are heavily used in the banking industry. New methods include positional information and incorporate image pixels from the document pages into the model (to be discussed in Section \ref{sec:methodology}).
The benefit of analyzing visually rich documents to improve business operations has been demonstrated in closely related industries such as insurance, where damage claim processes can be optimized \citep{levich2023}.
This example illustrates a transition towards a new era of multimodal document analysis, also highlighting the importance of leveraging diverse banking documents.

\paragraph{Research on Multimodal Models and Document Analytics in Banking}
In the field of multimodal models and document analytics for the banking sector, some pioneering studies have been carried out. \citet{engin2019} explore a multimodal deep neural network for the classification of Turkish banking documents using both textual and visual features. 
They categorize document layout structures into free-form text, large tables, custom forms, and forms predefined by specific organizations \citep{engin2019}. Furthermore, \citet{oral2020} demonstrate the power of deep learning algorithms and word position features in extracting information from banking documents, specifically money transfer orders. \citet{oral2022} take a more technical approach by examining the impact of different fusion techniques on the extraction of information from unstructured documents, again focusing on money transfer orders.

However, research on multimodal document analytics in banking remains limited, often focusing on specific document types. This leaves a large number of potential applications unexplored. A wider range of document types and applications for multimodal document analytics in banking is yet missing.
On a side note, even the most recent multimodal models, such as GPT-4, remain unexplored for their potential in banking document analytics.

Our study seeks to fill this gap by first reviewing the various types of banking documents. The goal is to build a bridge between the methodological literature and practical banking applications, a connection that is currently underrepresented. In particular, our objective is to assess the layout variability and automation potential of banking documents, aspects that are rarely addressed in previous works.
Subsequently, we propose a robust multimodal document analysis method using LayoutXLM on German company register extracts. To the best of our knowledge, this is the first study that leverages a pre-trained multimodal model to conduct document analytics in the banking sector. This includes a detailed evaluation to address challenges such as class imbalance, learning curve analyses, and component-wise ablations.
Through these efforts, our study aims not only to deepen the understanding of multimodal document analysis in banking, but also to pave the way for future research in this promising field.
\section{Methodology}
\label{sec:methodology}
The methodology section clarifies the technical foundations of the study. First, we discuss the general principles of multimodal models before addressing specific applications in banking. The second half of the section focuses on multimodal document analytics and concludes with an in-depth exploration of the LayoutXLM model.

Unimodal models such as the Transformer BERT \citep{devlinBert} or the GPT-3 series \citep{brown2020language}, have made significant strides in text analysis. However, their focus on a single data type can miss relevant information in industries with various data sources, such as banking \citep{zhu2015}.


To bridge this gap, there is a growing body of research on multimodal models that integrate different types of data to understand complex data. Early, intermediate and late fusion describe the integration point within the model and have an impact on overall performance \citep{boulahia2021, zhu2015}. Multimodal models tend to deliver superior accuracy over unimodal predecessors \citep{poria2017}.


Multimodal approaches already demonstrate promising results in banking. For example, \citet{lee2019} find that merging different sources of stock market data, using early or intermediate fusion, improves the accuracy of stock market predictions, while \citet{dang2018} also include textual data in their scope to effectively predict short-term stock trends. 
\citet{wang2023} implement multimodal financial statement fraud detection with textual and financial data. Similarly, \citet{stevenson2021} and \citet{tavakoli2023} use multimodal models for credit risk prediction, demonstrating their practical application in supporting decisions within banking.

\subsection{Multimodal Document Analytics}
\label{sec:multimodal-document-analytics}
Table \ref{tab:literature-methodology2} shows how the evolution of document analytics has been marked by the integration of multimodal data, extending the capabilities of traditional unimodal natural language understanding (NLU) models.

\begin{table}[htbp]
  \centering
  \caption{Methodological Overview of Document Analytics Frameworks}
  \label{tab:literature-methodology2}
  \setlength\tabcolsep{33pt}
  \renewcommand{\arraystretch}{1.2}
  \scriptsize
    \begin{tabularx}{\textwidth}{lccc}
    \toprule
    \textbf{Study} & \textbf{Presented Model} & \textbf{Multimodal} & \textbf{Cross-lingual} \\
    \midrule
    \addlinespace[1ex]
    \citet{devlinBert} & mBERT & & X \\
    \citet{conneauXLM} & XLM& & X \\
    \citet{conneauXLMRoberta} & XLM-RoBERTa & & X \\
    \citet{brown2020language} & GPT-3 & & X \\
    \citet{xuLayoutLMv1} & LayoutLM & X & \\
    \citet{xuLayoutlmv2} & LayoutLMv2 & X & \\
    \citet{liSelfDoc} & SelfDoc & X & \\
    \citet{chiInfoXLM} & InfoXLM & X & X \\
    \citet{xu2021layoutxlm} & LayoutXLM & X & X \\
    \citet{liMarkupLM} & MarkupLM & X & \\
    \citet{chenXDoc} & XDoc & X & \\
    \citet{huang2022layoutlmv3} & LayoutLMv3& X & \\
    \citet{liDIT} & DiT & X & (X)\\
    \bottomrule
    \end{tabularx}
\end{table}

Despite their proficiency in handling textual data, early NLU models such as mBERT \citep{devlinBert}, XLM \citep{conneauXLM}, XLM-RoBERTa \citep{conneauXLMRoberta}, and GPT-3 \citep{brown2020language} are bound by their unimodal structure, making them insufficient for handling multimodal data. The introduction of LayoutLM \citep{xuLayoutLMv1}, which jointly analyzes text and layout, marks a shift towards multimodal document analysis. Building on this development, LayoutLMv2 \citep{xuLayoutlmv2} offers a refined approach and significantly enhances the visual-rich document understanding capabilities (VrDU).
In parallel, SelfDoc \citep{liSelfDoc} employs a self-supervised pre-training framework, while MarkupLM \citep{liMarkupLM} presents a pre-trained model to analyze text and markup information from websites. However, the scope of research on multimodal models is mostly limited to English, with some exceptions such as the DiT model \citep{liDIT}, which is also trained on Chinese documents.

To address a more global applicability, the cross-lingual pre-training model, InfoXLM \citep{chiInfoXLM}, is introduced to fill this gap. This advancement becomes instrumental in the development of LayoutXLM \citep{xu2021layoutxlm}, a multimodal cross-lingual document analytics foundation model that enables VrDU in 53 different languages. Despite the notable progress in the development of GPT-4,  in particular its multimodal capabilities that focus on image understanding and image content querying, it still lacks the ability to understand the spatial and layout information of word tokens \citep{openai2023gpt4}. Therefore, it is not included in this study. 

\subsection{LayoutXLM}
LayoutXLM uniquely combines the benefits of multimodal learning and cross-lingual understanding.
Its approach positions it as a leading model in this field, making it a cornerstone of our study \citep{xu2021layoutxlm}.

\subsubsection{Architecture}
LayoutXLM adopts the transformer-based architecture of LayoutLMv2 \citep{xuLayoutlmv2}, which takes inputs from three different modalities: text, layout, and images. These inputs are encoded using text embedding, layout embedding, and visual embedding layers. 

Embeddings are essential to deep learning models as they transform data into structured representations that can express complex syntactic and semantic information. They provide a solution to the high dimensionality and sparsity of raw data by transforming the data into low-dimensional dense vectors, which facilitates efficient computation and storage \citep{mikolov2013efficient}.

\begin{figure}[h]
\centering
\includegraphics[trim=75 155 70 60, clip,width=1\textwidth]{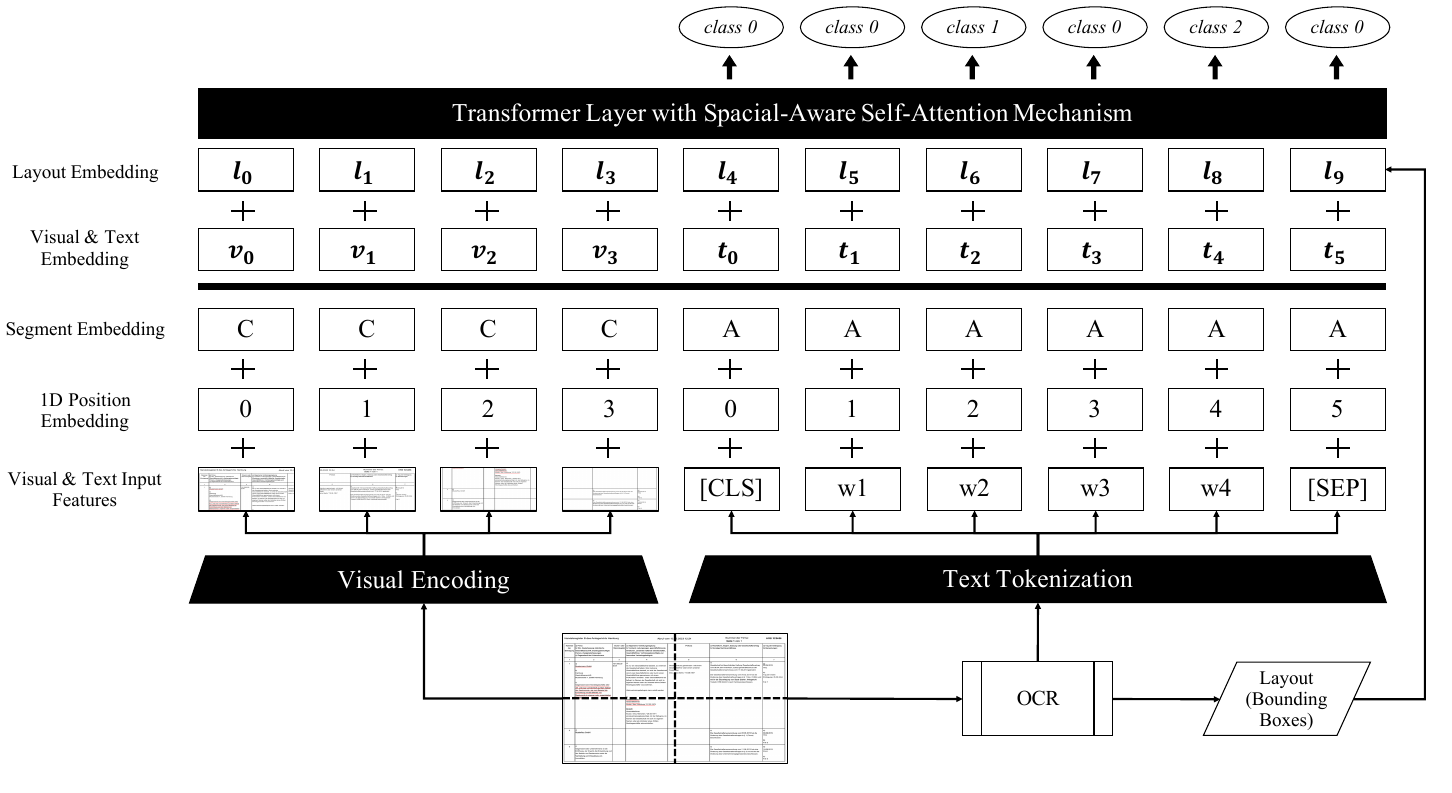}
\caption{Architecture of LayoutXLM in a Token Classification Setup.}
\label{Fig:LayoutXLM}
\end{figure}

Figure \ref{Fig:LayoutXLM} depicts the overall architecture of LayoutXLM \citep{xuLayoutlmv2, xu2021layoutxlm} in a token classification setup. In this section, we examine the individual components.

\paragraph{Text Embedding}
Essentially, each token's text embedding is derived from its own identity (the token itself), its position in the sequence, and its source segment.
To obtain the text embedding, the document's text is first tokenized into a sequence. This sequence begins with a [CLS] token, ends with a [SEP] token for each text segment, and is padded with [PAD] tokens to reach the maximum sequence length (L). Now, the $i$-th ($0 \leq i \le L$) text embedding is the sum of the following three components:

\begin{enumerate}
  \item The token embedding $TokEmb(w_i)$, which represents the token itself.
  \item The 1D positional embedding $PosEmb1D(i)$, which signifies the position of the token within the sequence.
  \item The segment embedding $SegEmb(s_i)$, which indicates the segment of the text from which the token originates.
\end{enumerate}

Thus, the $i$-th text embedding is represented as:
\[
t_i = \textrm{TokEmb}(w_i) + \textrm{PosEmb1D}(i) + \textrm{SegEmb}(s_i)
\]

\paragraph{Visual Embedding}

In the visual embedding step, each page image $I$ is resized to $224 \times 224$ and fed into a Convolutional Neural Network (CNN) to generate a fixed-length visual token embedding sequence. These embeddings are then unified in dimensionality using a linear projection $\textrm{Proj}(\textrm{VisTokEmb}(I)_i)$, ensuring consistency across different inputs. 
Since the CNN does not capture positional information, a shared 1D positional embedding $\textrm{PosEmb1D}(i)$, similar to the one used in the text embedding, is added to each visual token to provide positional information.  All visual tokens are grouped into the visual segment $C$. The final $i$-th visual embedding is expressed as:

\[
v_i = \textrm{Proj(VisTokEmb}(I)_i) + \textrm{PosEmb1D}(i) + \textrm{SegEmb}(C)
\]

where $0 \leq i < WH$. Dimensions $W$ and $H$ represent the size of the output feature map after processing the image through CNN, giving us a total of $WH$ visual tokens. This representation combines the visual features, positional information, and segment assignment for each visual token.

\paragraph{Layout Embedding}
The layout embedding encodes the spatial layout information derived from the OCR results, including data about the bounding box (width, height, and corner coordinates) of each token. These coordinates are normalized and discretized to integers in the range [0, 1000]. 

For the $i$-th text or visual token (with $0 \leq i < WH + L$), denoted by its normalized bounding box $box_i = (x_{min}, x_{max}, y_{min}, y_{max}, width, height)$, the layout embedding layer combines these six bounding box features to construct a 2D positional (layout) embedding $l_i$, with
\[
l_i = \textrm{Concat}(\textrm{PosEmb2D}_x(x_{min}, x_{max}, width), \\
\textrm{PosEmb2D}_y(y_{min}, y_{max}, height))
\]

The layout embedding hence encapsulates the spatial arrangement or layout information of the document. Special tokens such as [CLS], [SEP], and [PAD] are associated with an empty bounding box $(0, 0, 0, 0, 0, 0)$.

\paragraph{Multimodal Encoder with Spatial-Aware Self-Attention Mechanism}

The three embeddings - text, visual, and layout - are then combined using an early fusion approach, where the information from all the modalities is merged right from the beginning \citep{zhu2015, boulahia2021}. This combined multimodal input is fed into a transformer-based encoder, allowing the model to learn a joint representation where all modalities can influence each other during the encoding process.
Visual embeddings ${v_0, ..., v_{WH-1}}$ and text embeddings ${t_0, ..., t_{L-1}}$ are concatenated into a unified sequence $X = {v_0, ..., v_{WH-1}, t_0, ..., t_{L-1}}$. Spatial information is integrated into this sequence by adding layout embeddings to each corresponding element, yielding the $i$-th (with $0 \leq i < WH + L$) first-layer input $x^{(0)}_i = X_i + l_i$.


The core of the LayoutXLM encoder follows the transformer architecture, consisting of a stack of multi-head self-attention layers followed by a feed-forward network. However, unlike the original self-attention mechanism, LayoutXLM uses a spatially-aware self-attention mechanism to incorporate relative position information and efficiently model local invariance in the document layout \citep{xuLayoutlmv2}. Its unified approach of combining text, visual and layout information allows LayoutXLM to capture the rich information present in document images and facilitates powerful downstream applications such as document understanding and information extraction \citep{xu2021layoutxlm}.

\subsubsection{Pre-Training and Fine-Tuning}
The pre-training objectives of LayoutXLM are adapted from the LayoutLMv2 framework, which has shown efficacy in VrDU.
LayoutXLM undergoes pre-training with three tasks, Multilingual Masked Visual-Language Modeling, Text-Image Alignment, and Text-Image Matching, which together strengthen the model's understanding of language, visuals, and spatial relationships in documents \citep{xuLayoutlmv2}.

The pre-training tasks enable LayoutXLM to learn an understanding of documents at a semantic and visual level, as well as layout and spatial information. The extensive pre-training process allows LayoutXLM to use transfer learning, where the pre-trained model can be effectively tuned for specific tasks even with smaller labeled datasets. This adaptability makes LayoutXLM a versatile foundation model for document analytics that can be used in a wide range of downstream tasks. Its value is particularly pronounced in the banking sector, where document analysis is a key success factor. Integrating LayoutXLM into the workflow of a bank can enhance its processes and drive efficiency and decision-making.

\section{Documents in Banking}
\label{sec:bankingdocs}

Beyond the technical aspects of foundational document analytics models, practical applications with high business value are crucial. Bank documents, distributed in multiple languages and layouts, align well with the multimodal, cross-lingual capabilities of LayoutXLM. To fully understand the potential of document analytics in banking, we now examine the complexity of its document landscape.

\subsection{Structure of Commercial Banking Business}
\label{sec:structure-banking}
\begin{figure}[h]
\centering
\begin{forest}
for tree={
    grow'=east,
    draw,
    align=center,
    parent anchor=east,
    child anchor=west,
    edge path={\noexpand\path[\forestoption{edge}](\forestOve{\forestove{@parent}}{name}.parent anchor)--++(0pt,0)--(\forestove{name}.child anchor)\forestoption{edge label};},
    font=\scriptsize,
    l sep+=20pt, 
    s sep+=5pt, 
    align=right
}
[Banks
    [Core Business
        [Customer Business
            [Client Relationship Management]
            [Account and Transaction Services]
            [Investment and Savings Solutions]
            [Credit and Lending Services]
            [Trade and Guarantee Services]
        ]
        [Proprietary Trading, edge=dashed, dashed]
    ]
    [Internal Operations, edge=dashed, dashed]]
]
\end{forest}
\vspace{5px}
\caption{Overview of the Business Structure of a typical Commercial Bank.}
\label{fig:banking-overview}
\end{figure}
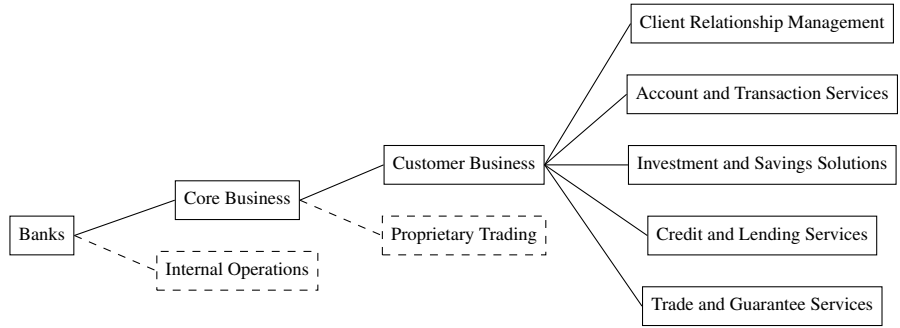

The general structure of the commercial banking business is illustrated in Figure \ref{fig:banking-overview}. This diagram provides an overview of the key areas within a typical bank. Banking activities are first divided into the core business and other internal operations, which are not directly related to banking-specific departments (e.g., human resources).

Although internal operations and proprietary trading are integral components of banks' activities, they are outside the scope of this study. The focus is on customer-centric core operations. While other internal operations within banks are indeed document intensive, the specificity of focusing on customer-facing processes provides a deeper understanding of the unique characteristics of the banking industry. This approach differentiates our study from more general document analytics research and allows us to directly impact process efficiency, risk management, and customer-related studies.

This study explores the core business branch of the tree, specifically the customer business, which can be further subdivided into typical banking areas. The proposed classification for these areas in banking provides a framework for understanding the activities of a commercial bank.
\citet{casu2015introduction} highlight typical components of the core customer business, including payment services, deposits, investments, lending services, e-banking, and additional financial services (e.g., insurance or trade finance). To account for contextual variations in banking activities and subprocesses, and to better align with our research focus, we propose a refined categorization:

\begin{enumerate}
    \item \textbf{Client Relationship Management.} In this domain, the focus lies on managing the bank-customer relationship, including the know-your-customer process and general customer requests.
    
    \item \textbf{Account and Transaction Services.} This segment includes the management of checking accounts and the facilitation of payment transactions. 
    
    \item \textbf{Investment and Savings Solutions.} Within this area, a diverse range of savings accounts and investment services are offered, including stock trading.
    
    \item \textbf{Credit and Lending Services.} This category encapsulates all activities related to the provision of loans and other forms of credit business. For corporate clients, this also includes specialized financing solutions, such as factoring and leasing.
    
    \item \textbf{Trade and Guarantee Services.} This area covers additional or specialized financial services such as trade finance and guarantee business, which have unique document management requirements.
\end{enumerate}

Some banks may offer additional financial and non-financial services (e.g. insurance). Our classification does not consider such services separately because they can either be embedded in one of the areas presented or fall outside the scope of core banking services.

Our contextual framework provides a clear perspective on the core business activities of commercial banks, particularly with regard to document analytics. Each area represents a distinct domain of document-intensive processes. The framework facilitates a focused exploration of the role and potential of multimodal document analytics within core banking services and serves as the guiding principle for our investigation.

\subsection{Assessment and Categorization of Banking Documents}
In order to effectively assess the characteristics of the banking documents in the areas presented, we employ certain assessment and categorization criteria. The aim is to provide a comprehensive overview to improve the understanding of each type of document and its potential for process automation.

In the course of commercial core banking, each operation or process typically requires specific documents. We have undertaken a collection of various types of documents to cover a wide range of banking areas. For example, the consumer loan approval process may involve documents such as income proofs and credit ratings.
Given the variety of documents in each area of commercial banking, we conduct an evaluation for each type of document. A systematic review of relevant literature, process documentation, practice guides, web resources, and consultations with banking domain experts provides a solid basis for this assessment. Each document type is evaluated against three key criteria:

\begin{itemize}

\item \textbf{Multimodality Assessment.} As proposed by \citet{engin2019}, different structures within the documents range from free-formatted texts to highly standardized forms. In our assessment, we consider the presence of unstructured text, tabular data or forms, and figures or images. The reason for this diversity consideration is its impact on our approach to document analytics. For example, documents that primarily consist of free text might only require the application of unimodal NLP models, and documents that are composed of images may be best served by traditional unimodal image classifiers. However, those with a mixture of text, forms, and images may require multimodal methods such as LayoutXLM.

\item \textbf{Automation Potential.} Each document is analyzed according to its potential for automation, indicating the expected impact and the ease of automating its processing. This assessment accounts for the current manual effort required to process each document and the possible benefits that could arise from automation, such as information extraction. Starting document analytics with use cases that have high automation potential can ensure efficient resource allocation and maximum return on investment.

\item \textbf{Layout Variability.} This criterion indicates how much the appearance of a particular document type can vary. If a document type always follows a strict format and appearance, its layout variability is considered low, making it easier to process. On the contrary, a document type that appears in multiple forms has a high layout variability, potentially making its processing more challenging.
The assessment of layout variability is based on an examination of multiple examples of each document type. 

\end{itemize}

Lastly, while the additional assessment of \textit{data privacy} could have been a separate criterion, we observe that almost all banking documents contain sensitive information resulting in strict privacy and security measures. Therefore, we emphasize the importance of data privacy as a critical consideration for all types of documents. However, some types of documents, such as company register extracts, may be public and less sensitive. 

These assessment criteria provide a framework for understanding the current landscape of banking documents and the opportunities they present for automation and decision support. The proposed framework lays a foundation for the effective adoption of document analytics in commercial banking operations.

\subsection{Results and Applications of Document Analytics in Banking}
Table \ref{tab:doc_assessment} provides an overview of various document-intensive banking processes and the corresponding assessment of multimodality, automation potential, and layout variability.

\begin{landscape}
\begin{table}[h]
\centering
\caption{Assessment of Banking Documents}
\label{tab:doc_assessment}
\setlength\tabcolsep{5pt} 
\begin{scriptsize} 
\begin{tabular}{lp{5.5cm}ccccc}
\toprule
\textbf{Area} & \textbf{Processes and Documents} & \multicolumn{3}{c}{\textbf{Multimodality}} & \textbf{Automation} & \textbf{Layout} \\
 & & \textit{Unstructured} & \textit{Tabular} & \textit{Figures or} & \textbf{Potential} & \textbf{Variability} \\
 & & \textit{Texts} & \textit{Data/Forms} & \textit{Images} & & \\
\midrule
\addlinespace[1ex]

\multirow{11}{*}{Client Relationship Management} & \textbf{Onboarding Process} &  &  & &  &  \\
 & \hspace{1em}\textit{Identification Documents} & &  &  &  &  \\
 & \hspace{2em}ID Card / Passport & & X & X & High & Low \\
 & \hspace{2em}Proof of Address & & X &  & High & Low \\
 & \hspace{2em}Company Register Extract & X & X & (X) & High & Low \\
& \hspace{2em}Articles of Association & X & & & High & Medium \\
& \hspace{2em}Organizational Chart & X & & X & High & High \\
 & \hspace{1em}KYC Form & & X & & High & Low \\
\addlinespace[1ex]
 & \textbf{Customer Service} &  &  & &  &  \\
 & \hspace{1em}Textual Service Requests/Complaints & X & & & High & High \\
 & \hspace{1em}Product Orders or Cancellation & X & (X) & & High & High \\
\midrule
\addlinespace[1ex]

\multirow{19}{*}{Account and Transaction Services} & \textbf{Account Opening Process} &  &  & &  &  \\
 & \hspace{1em}Account Opening Contract & X & X & & Medium & Medium \\
 & \hspace{1em}Authorization Form & & X & & Medium & Medium \\
 & \hspace{1em}Signature Specimen & & & X & Low & High \\
 \addlinespace[1ex]
 & \textbf{Transaction Process} &  &  & &  &  \\
 & \hspace{1em}Standard Payment Form & & X & & High & Low \\
 & \hspace{1em}Informal Payment Order & X & & & Low & High \\
\addlinespace[1ex]
 & \textbf{Account Closure Process} &  &  & &  &  \\
 & \hspace{1em}Account Closure Form & & X & & Medium & Low \\
 & \hspace{1em}Certificates of Inheritance & X & X & & Medium & High \\
 & \hspace{1em}Switch Bank Orders & X & X & & High & Medium \\
\addlinespace[1ex]
 & \textbf{Others} &  &  & &  &  \\
 & \hspace{1em}Account Statement & & X & & High & Medium \\
 & \hspace{1em}Balance Confirmation & & X & & High & Medium \\
 & \hspace{1em}Garnishment Order & X & X & & High & Medium \\
  & \hspace{1em}Invoices & & X & & High & High \\
\midrule
\addlinespace[1ex]

\multirow{11}{*}{Investment and Savings Solutions} & \textbf{Standard Savings Process} &  &  & &  &  \\
 & \hspace{1em}Savings Certificates / Passbook & & X & & Medium & Medium \\
 & \hspace{1em}Deposit Slip & & X & & Medium & Medium \\
\addlinespace[1ex]
 & \textbf{Investment Process} &  &  & &  &  \\
 & \hspace{1em}Investment Application Form & & X & & Medium & High   \\
 & \hspace{1em}Investment Advice Record & X & X & & Medium & High \\
 & \hspace{1em}Transaction Receipt & & X & & Medium & Medium \\
 & \hspace{1em}Securities Account Statements & & X & & High & Medium \\
 & \hspace{1em}Effective Securities & & & X & Low & High \\
\addlinespace[1ex]
 & \textbf{Tax Documents} &  &  & &  &  \\
 & \hspace{1em}Exemption Form & & X & & High & Low \\
\bottomrule
\end{tabular}
\end{scriptsize} 
\end{table}
\end{landscape}

\begin{landscape}
\begin{table}[h]
\centering
\caption*{Table \ref{tab:doc_assessment} (continued): Assessment of Banking Documents}
\label{tab:doc_assessment2}
\setlength\tabcolsep{5pt} 
\begin{scriptsize} 
\begin{tabular}{lp{5.5cm}ccccc}
\toprule
\textbf{Area} & \textbf{Processes and Documents} & \multicolumn{3}{c}{\textbf{Multimodality}} & \textbf{Automation} & \textbf{Layout} \\
 & & \textit{Unstructured} & \textit{Tabular} & \textit{Figures or} & \textbf{Potential} & \textbf{Variability} \\
 & & \textit{Texts} & \textit{Data/Forms} & \textit{Images} & & \\
\midrule
\addlinespace[1ex]

\multirow{19}{*}{Credit and Lending Services} & \textbf{Loan Approval Process} &  &  & &  &  \\
 & \hspace{1em}Loan Application Form & (X) & X & & High & Medium \\
 & \hspace{1em}\textit{Income Proof} &  & & &  &  \\
 & \hspace{2em}Salary Statement & (X) & X & & High & Medium \\
 & \hspace{2em}Pension Statement & X & X & & High & Medium \\
 & \hspace{2em}Financial Status Report & X & X & & High & High \\
 & \hspace{2em}Annual Company Reports & X & X & X & High & High \\
 & \hspace{1em}\textit{Credit Ratings} &  & & &  &  \\
 & \hspace{2em}Rating Form & X & X & & Medium & Medium \\
 & \hspace{2em}Third-Party Ratings & X & X & & High & High \\
 & \hspace{1em}Bank Reference & X & X & & High & High \\
 & \hspace{1em}\textit{Loan Agreement} &  & & &  &  \\
 & \hspace{2em}Loan Agreement Contract& X & X & & Medium & Medium \\
 & \hspace{2em}Loan Securities Documents & X & X & X & High & High \\
 & \hspace{2em}Loan Guarantee & X & X & & Medium & High \\
 & \hspace{2em}Land Register Extracts & X & X & (X) & High & Low \\
 & \hspace{1em}\textit{Loan Servicing} &  & & &  &  \\
 & \hspace{2em}Loan Repayment Schedule & X & X & & High & High \\
\addlinespace[1ex]
 & \textbf{Others} &  &  & &  &  \\
 & \hspace{1em}Factoring / Leasing Invoices & X & X & & High & High \\
 & \hspace{1em}Dunning Process Documents & X & X & & Low & Medium \\
 & \hspace{1em}Insolvency Documents & X & X & & Low & High \\
\midrule
\addlinespace[1ex]

\multirow{16}{*}{Trade and Guarantee Services} & \textbf{Trade Finance Application Process} &  &  & &  &  \\
 & \hspace{1em}Trade Finance Application Form & X & X & & High & Low \\
 & \hspace{1em}Trade Agreement & X & & & Medium & Medium \\
 & \hspace{1em}Commercial Invoices & X & X & & High & High \\
\addlinespace[1ex]
 & \textbf{Document Verification Process} &  &  & &  &  \\
 & \hspace{1em}Trading Documents & X & X & & High & High \\
 & \hspace{1em}Transportation Documents & X & X & & Medium & High \\
 & \hspace{1em}Insurance Documents & X & X & & Medium & High \\
 & \hspace{1em}Customs Documents & X & & & Medium & Medium \\
 & \hspace{1em}Additional Certificates & X & X & (X) & Medium & High \\
\addlinespace[1ex]
 & \textbf{Guarantee Issuance Process} &  &  & &  &  \\
 & \hspace{1em}Guarantee Document & X & X & & High & High \\
 & \hspace{1em}Letter of Credits & X & X & & High & High \\
 & \hspace{1em}Export Guarantees & X & X & & High & High \\
 & \hspace{1em}Other Guarantees & X & X & & Medium & High \\
 & \hspace{1em}Rental Payment Guarantee & X & X & & Medium & Medium \\
 & \hspace{1em}Delivery Guarantee & X & X & & Medium & Medium \\
\bottomrule
\addlinespace[0.5ex]
\multicolumn{7}{p{\linewidth}}{Note: Documents can be used in multiple processes. For this study, each document is assigned to one primary process based on common usage.} 

\end{tabular}
\end{scriptsize} 
\end{table}
\end{landscape}

These processes are covered by the five customer-centric core banking areas that we previously defined in \ref{sec:structure-banking}. It should be noted that this presentation is not exhaustive and is meant as a representative example of common processes in banks.

An important insight that emerges from Table \ref{tab:doc_assessment} concerns the inherent multimodality of banking documents. Many documents not only consist of plain text but also include various types of visual content, such as images, graphs, tables, or even handwriting. This complexity necessitates the application of multimodal models like LayoutXLM to ensure a comprehensive understanding of the document's content. Forms such as those for loan applications or account openings exemplify this characteristic. They contain tabular data, to be completed by the customer, and unstructured text. Likewise, identification documents, which are required for know-your-customer purposes, typically include personal information presented in (tabular) text format, along with photographs and signatures. In both cases, an unimodal approach that focuses solely on textual or visual data would be insufficient.

Recognizing this challenge, we take a closer look at how banks can optimally harness the potential of multimodal documents. To do this, we've identified specific use cases that not only leverage this multimodal nature but also provide concrete solutions to common banking needs. In particular, we identify three key applications: information extraction, document page classification, and document splitting, each of which has the potential to improve the efficiency of banking operations.

\paragraph{Information Extraction}
An advanced use case is the automated extraction of structured information from unstructured documents, including elements such as entity names, numeric data, and checkbox selections. For example, extracting customer details from ID documents or company information from company register extracts or annual company reports. Accurate extraction of these data enables banks to process applications or requests quickly and efficiently, reducing processing times and improving customer satisfaction.
The complexity of information extraction is closely related to the layout variability of the document type. For example, highly standardized forms, such as tax forms and identity documents, are generally easier to parse and extract information.

\paragraph{Document Page Classification}
This involves categorizing document pages into predefined document types. This is especially important when banking processes require different documents, each of which must be processed differently. By automatically identifying and then processing each type of document, banks can increase operational efficiency and ensure consistent and accurate document lifecycle management. 
This approach can be particularly beneficial for document-intensive processes such as opening an account. In this case, specific documents such as account opening forms, identity proofs, and other supporting documents can be automatically classified and routed to streamline the process.

\paragraph{Document Splitting}
Document splitting refers to the process of breaking multipage documents into smaller, more manageable segments or pages, where each segment represents a distinct sub-document. 
These sections may correspond to predefined document types, or they may represent individual units of information within a document type. For example, a compilation of monthly salary proof documents can be divided not only based on their categorization as ``salary proof", but also according to the specific month to which they correspond.
This process is, for example, required when a batch of physical documents is scanned and all pages are combined into one digital file. In this context, document splitting helps to separate and organize the collected information into individual units, making it easier to handle and process.
\\

These use cases demonstrate how document analytics can serve as a powerful tool in the banking sector. It can automate manual processes, increase accuracy, and accelerate decision-making, contributing to increased customer satisfaction and competitive advantage. Other potential applications could include document representation, visual question answering, and more, expanding the range of possibilities in this area.

\section{Experimental Design}
\label{sec:experimental}

In this section, we explain our experimental framework. We start with the motivation for our chosen use case, which involves the extraction of information from German company register extracts. These documents present textual data within a tabular structure, often incorporating historical information indicated by red text or underlining. Therefore, the chosen document type might benefit from a model that interprets not only textual data but also visual cues. The use of German data introduces an added level of complexity as it requires a model with foreign language capabilities, reflecting the common needs in global banking processes.

One particular challenge associated with Optical Character Recognition (OCR) on tabular data is its row-by-row reading pattern. Although the semantic token structure might be column-based in tabular data, tokens are read sequentially across the rows, which could potentially misalign the semantic interpretation when processing scans because the reading direction crosses the table columns.

The aim is to classify each word token into one of the following classes: company name, legal form, headquarters, capital number, capital currency, director (geschaeftsfuehrer), authorized officer (prokurist), limited partner (kommanditist), and shareholder (gesellschafter). Tokens that do not fit into any of these categories are called ``other", resulting in a total of 10 classes. A notable aspect of this token classification task is the significant class imbalance, with about 95\% of tokens falling into the ``other" category. This presents a further challenge for the model's learning and performance optimization.

\begin{figure}[h]
\centering
\includegraphics[width=1\textwidth]{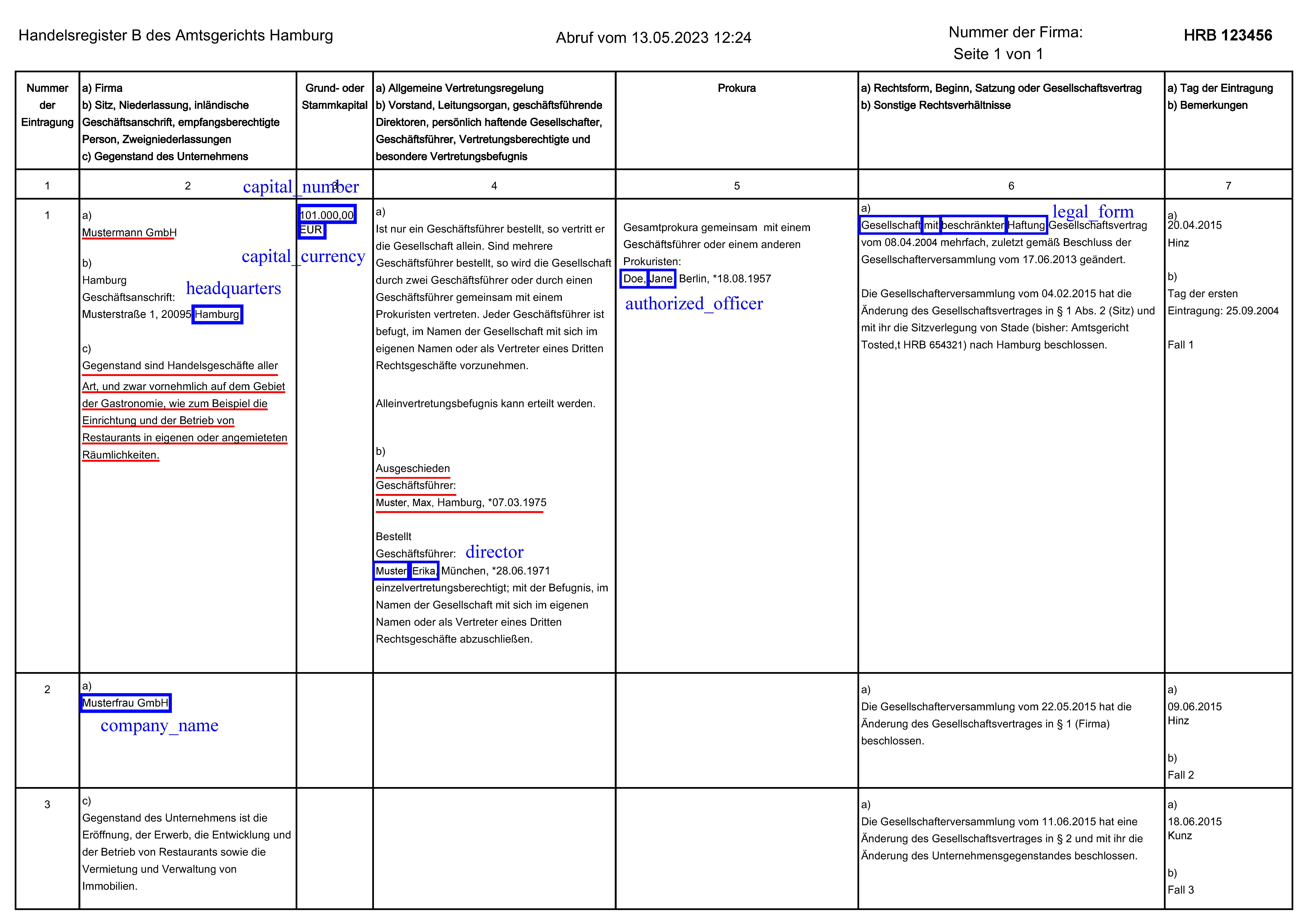}
\caption{Company Register Extract with Labels from fine-tuned LayoutXLM Model.}
\label{Fig:cr-labels}
\end{figure}

Figure \ref{Fig:cr-labels} exemplifies the typical structure of extracts from the German company register. Red underlining marks historical information, while blue boxes indicate the token labels (only tokens that do not belong to the ``other" class are visualized). 

Next, we describe our data collection and processing, detail model selection, and outline evaluation and benchmarking approaches.

\paragraph{Data Collection and Processing}
We collect extracts from the public German company register for more than 500 companies in different industries. This gives a total data set of 2441 PDF pages. Additionally, we use company metadata XML files to derive labeling information in a structured format. We then iterate over the OCR document tokens and perform a metadata-based token tagging process. This matching process is refined by analyzing the color of the text or its underlining and removing any entity labels from historical information (indicated by red highlighting). Finally, we exclude empty pages and retain only those pages containing at least three token labels with relevant company entity information, resulting in a data set of 1503 pages.

\paragraph{Model Fine-Tuning and Benchmarking}

We chose the pre-trained LayoutXLM model from Huggingface for the token classification task.
To optimize its performance for our specific use case, we put the model through additional training epochs beyond its pre-trained state. This fine-tuning is specifically aimed at the task of extracting information from company register extracts.

The evaluation relies on a hold-out validation set that comprises 30\% of the data, or approximately 480 pages. This set is used in all ablation studies and comparisons to ensure the comparability and consistency of the results. We benchmark the performance of LayoutXLM against a BERT multilingual model and GPT-3.5 Turbo, which we access through its API using the F1 score as a performance metric.
For the GPT benchmark, we use the following prompt:

\begin{mdframed}[roundcorner=5pt,leftmargin=0cm,rightmargin=0cm,
  skipabove=1cm,skipbelow=1cm,
  frametitle=GPT Prompt,frametitlerule=true]
  \scriptsize
\texttt{Below is a list of word tokens extracted from a German company register extract scans. Please classify \\
each word token into one of the following classes: "company\_name", "legal\_form" (of the company), \\
"headquarters" (of the company), "capital\_number", "capital\_currency", "director\_(geschaeftsfuehrer)" \\
(only natural person name), "authorized\_officer\_(prokurist)" (only natural person name), \\
"limited\_partner\_(kommanditist)" (only natural person name), and "shareholder\_(gesellschafter)" (only \\
natural person name). For word tokens that do not fit into any of these categories, please label them \\
as "other". Also, if any
information is historic or outdated, assign the "other" class as well.
Your task is to create a full list of labeled tokens. Each labeled token should be in the format ["token", "class"]. The final output should be a list of the same length as the original list, with each token labeled \\ according to the appropriate class.
Note that the order of the tokens should remain the same as in the \\
original list. Here is the list of tokens for your consideration:\\
---}
\end{mdframed}
\section{Emperical Results}
\label{sec:results}

In this section, we present the empirical findings from our study centered on information extraction from German company register extracts, demonstrating the application of the LayoutXLM model on a non-English document. We fine-tune for this specific task, analyze the performance, and address the token class imbalance. This section further discusses metrics such as precision, recall, F1 scores, and overall accuracy on the validation data set, allowing a deeper understanding of potential banking applications.

In addition, an ablation study measures the contribution of each component of the model (text, layout, and image data) to its overall performance and explores the effect of different training sample sizes. 

As part of the analysis, we investigate the number of epochs required to fine-tune LayoutXLM. This is crucial to our understanding of the optimal balance between computational efficiency and performance outcome. By examining these empirical results, we aim to build a picture of how models like LayoutXLM can be used for document analysis tasks in banking.

\subsection{Performance Analysis of the LayoutXLM Model}

Examining the performance of the LayoutXLM model on our token classification task is the focus of this section. Due to the complexity of the task, zero-shot learning is not feasible, and fine-tuning the model is necessary. A full learning curve analysis for better understanding will be presented in the ablation studies (\ref{sec:ablation-studies}). Table \ref{tab:results-epoch10} shows the validation results obtained after fine-tuning a pre-trained LayoutXLM model for 10 epochs. This model achieved an overall accuracy of 0.9758 on the validation dataset. 
A closer analysis of these results reveals several findings.

\begin{table}[h!]
\setlength\tabcolsep{18.5pt} 
\centering
\small
\caption{Token Classification Results on Validation Set}
\label{tab:results-epoch10}
\begin{tabular}{lcccr}
\toprule
\textbf{Category} & \textbf{Precision} & \textbf{Recall} & \textbf{F1 score} & \textbf{Support}\\
\midrule
authorized officer & 0.6717 & 0.8840 & 0.7634 & 1000\\
capital currency & 0.7935 & 0.8488 & 0.8202 & 86\\
capital number & 0.7536 & 0.6265 & 0.6842 & 83\\
company name & 0.8105 & 0.9245 & 0.8637 & 384\\
director & 0.7314 & 0.6059 & 0.6628 & 373\\
headquarters & 0.8246 & 0.8910 & 0.8565 & 211\\
legal form & 0.9598 & 0.9728 & 0.9663 & 368\\
limited partner & 0.6026 & 0.7344 & 0.6620 & 64\\
shareholder & 0.7143 & 0.3704 & 0.4878 & 81\\
other & 0.9906 & 0.9841 & 0.9873 & 44927\\
\midrule
\textbf{Macro Average} & 0.7852 & 0.7842 & 0.7754 & 47577\\
\textbf{Weighted Average} & 0.9776 & 0.9758 & 0.9762 & 47577\\
\bottomrule
\end{tabular}
\end{table}

First, despite the severe token class imbalance present in the dataset, the LayoutXLM model proves to be robust, as it demonstrates considerable learning capabilities across all classes. This becomes evident from the F1 scores achieved by each class, even those with a smaller number of examples (lower support). For example, the class ``limited partner" has a support of only 64 and achieves an F1 score of 0.6620, indicating reasonable performance despite being underrepresented.

Second, the trade-off between precision and recall is evident in our results. For some categories, such as ``capital number", the model favors precision (0.7536) over recall (0.6265). This means that, while the model is cautious in its predictions for this class (leading to fewer false positives), it is missing a significant portion of true positives, leading to lower recall. On the other hand, for categories such as ``authorized officer", the model has a higher recall (0.8840) compared to its precision (0.6717), indicating that the model is able to identify a large portion of actual positives, but at the cost of including some false positives. Depending on the specific use case requirements, one may need to adjust the model or the decision threshold to optimize for either precision or recall.

These results underline the versatility of the LayoutXLM model in handling various document classes, exhibiting robustness to class imbalance, while maintaining reasonable precision and recall trade-offs in a banking context. The robustness of the model, as demonstrated by the macro-average F1 score of 0.7754, indicates an overall balanced performance across all classes despite the large support of the ``other" category. The choice of the macro-average F1 score, which calculates the average over all independent class scores, provides an assessment without favoring larger classes.

To provide a deeper understanding of how the LayoutXLM model performs in practice, the company register extract of Figure \ref{Fig:cr-labels} already includes a visualization of the output token labels. These are assigned by the model and can be mapped to the document.

\subsection{Ablation Studies}
\label{sec:ablation-studies}
The ablation studies dissect the LayoutXLM model to better understand its behavior and performance and shed light on when the model is effective in banking applications. To achieve this, we evaluate the impact of multimodal components and examine the effect of varying the size of the training data.

\subsubsection{Assessing the Contributions of Text, Layout, and Image Data}
\begin{figure}[h!]
\centering
\includegraphics[trim=35 5 55 37, clip,width=1\textwidth]{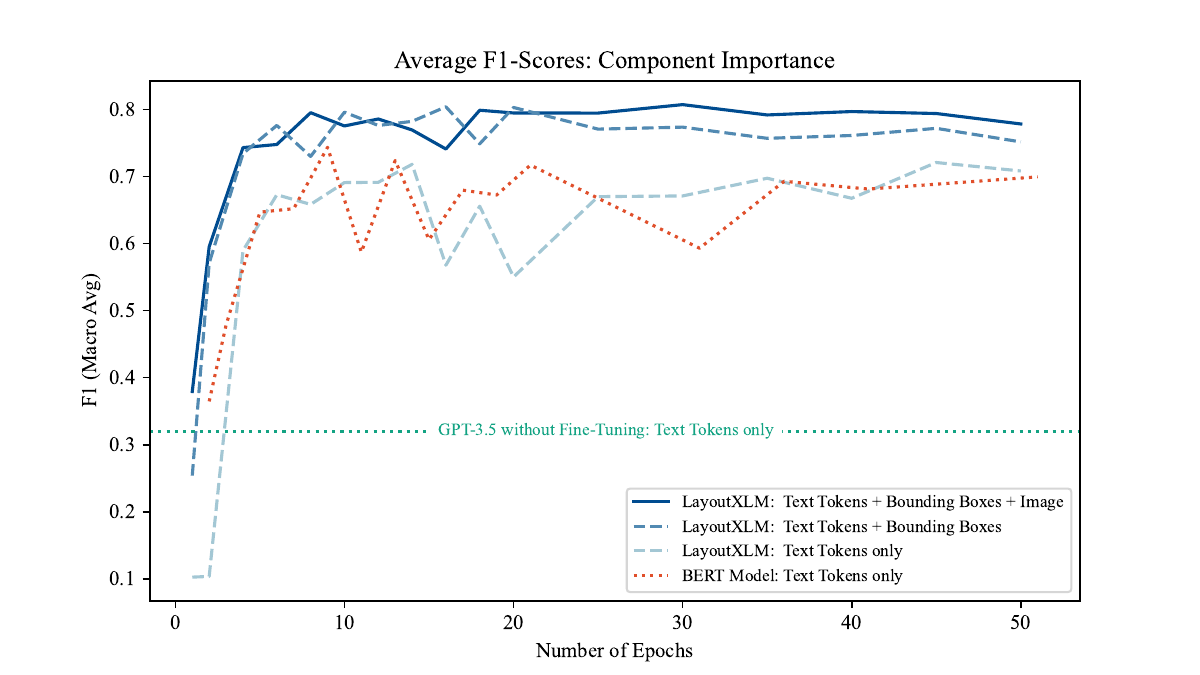}
\caption{Average F1-Score Comparison - Importance of Components.}
\label{Fig:component-importance}
\end{figure}

In the first part of the ablation studies, we analyze the contribution of different components, specifically text, layout information in the form of bounding boxes, and image data, in driving the performance of the token classification model. This analysis is illustrated in Figure \ref{Fig:component-importance}, which presents the learning curves initiated from the first epoch of fine-tuning on the company register extracts. To obfuscate layout and image information, we use empty bounding boxes $(0,0,0,0)$ and white pixels, respectively. We also include a multilingual BERT model as a baseline comparison.

In German company register extracts, historical entries are typically highlighted with red or underlined text. Including image data thus allows the model to recognize these visually marked distinctions.
The learning curves depict rapid performance increases until reaching a plateau and convergence after approximately 10 epochs. The text-only models have a slightly delayed convergence, which begins at about 25 epochs.
The LayoutXLM model, which incorporates all three data types - text, layout, and image - shows the steepest learning curve and achieves the highest F1 scores of approx. 80\% after convergence. When the model input is reduced by obfuscating the image data, the F1 score decreases. Eliminating the positional layout data causes an even steeper drop in performance and further delays the learning progress, underscoring the crucial role of these three data inputs.

A comparison of LayoutXLM with a multilingual BERT model and GPT-3 supports our initial assumption. During fine-tuning, BERT, which relies solely on textual data, performs similarly to the text-only variant of LayoutXLM. When comparing the performance to GPT-3 as an off-the-shelf solution, a fine-tuned LayoutXLM model again shows better results. Although GPT-3 is able to make predictions across all classes and thus handle the token class imbalance, it only achieves an F1 score of 0.32 (Figure \ref{Fig:component-importance}). This confirms the efficacy of a fine-tuned multimodal LayoutXLM model in document analytics, which combines text, layout, and image data to achieve superior performance.

Another important takeaway from this study is the rapid performance improvement of LayoutXLM within the initial 5 epochs. This suggests a time-efficient fine-tuning of the model, which is beneficial for practical applications.

\subsubsection{Impact of Training Data Size on LayoutXLM Performance}

\begin{figure}[h]
\centering
\includegraphics[trim=35 0 55 37, clip,width=1\textwidth]{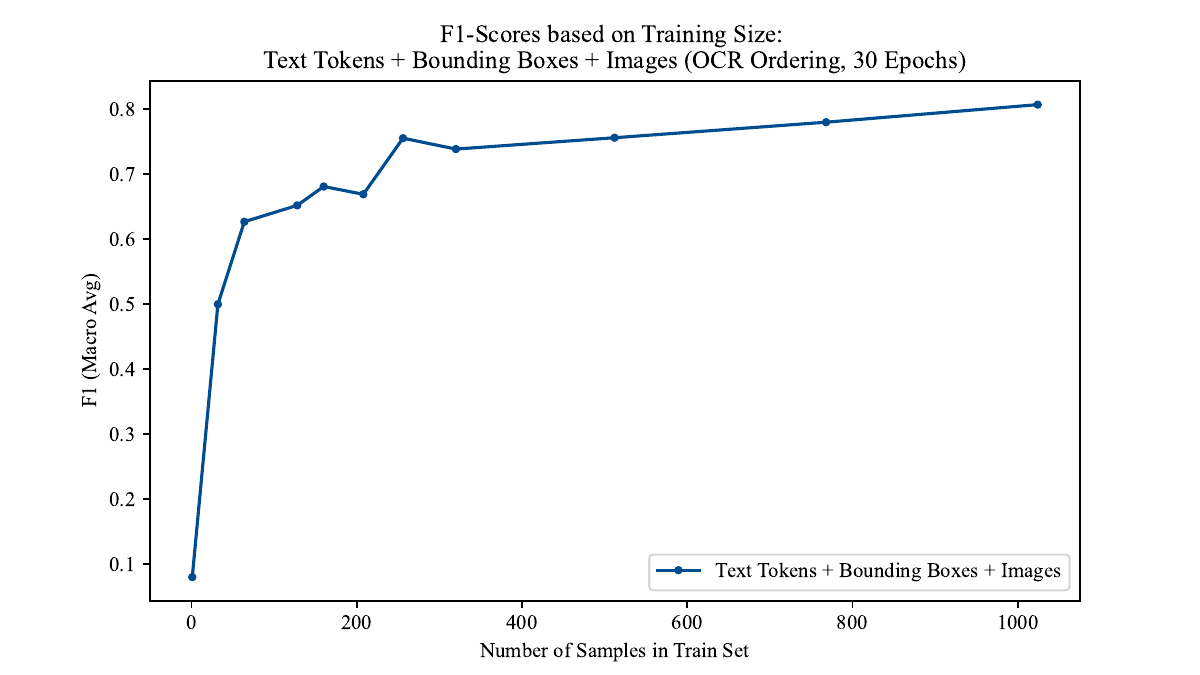}
\caption{Impact of Training Size on Validation Performance (30 Epochs Fine-Tuning).}
\label{Fig:train-size}
\end{figure}

To evaluate the effect of training data size, we subsample our data into smaller subsets, as illustrated in Figure \ref{Fig:train-size}. We fine-tune each subset over 30 epochs, taking into account the learning curve analysis, which indicated post-convergence at this point.

LayoutXLM exhibits considerable learning efficiency even with a minimal amount of labeled data. In particular, only 200-300 observations - a fraction of our full dataset - are sufficient for the model to effectively distinguish between the 10 token classes. This initial learning phase is characterized by a significant increase in the F1 score, which reaches 70\%. The performance gains slow down as the size of the dataset exceeds the 200-300 page mark and reaches a maximum of roughly 80\% when using the full training set. This observation underscores LayoutXLM's ability to use relatively small datasets for robust model training in practical settings.

The demonstrated efficiency of LayoutXLM in learning from limited labeled data has implications for organizations with limited resources for collecting and labeling data. It is possible to achieve respectable levels of performance with less data, although a larger dataset may still be beneficial for optimizing performance and capturing rare classes.

\section{Conclusion and Discussion}
\label{sec:conclusion}

The study considered the diverse landscape of banking documents, seeking opportunities to improve process efficiency. In this scope, we presented an overview of promising applications that can benefit from advanced document analytics. We further illustrated the application of the multimodal, cross-lingual LayoutXLM model for information extraction from German company register extracts. Empirical results confirmed the ability of LayoutXLM to handle visually rich banking documents. The ablation analysis further confirmed that the high performance is due to the model's ability to extract relevant information from layout and visual information, which underscores the importance of taking a multimodal approach toward document analytics. For example, this architecture outperforms established unimodal models such as BERT and GPT-3. LayoutXLM also demonstrated robustness against the inherent token class imbalance in company register extracts and delivered good results with little training data, supporting its suitability for banking applications.

Based on our study, we now discuss the implications for research and banking practice. First, our analysis and prototypical implementation highlight that banking documents display much variety in terms of style and content. Thus, we propose that analytical models in this area should also be adaptive and flexible to accommodate the dynamics inherent in banking documents.
More specifically, our exploration of the banking document landscape underscores their inherent multimodality. Many banking documents are a complex mixture of text, tables, graphics, and even handwriting. Component-by-component analysis of LayoutXLM clearly demonstrates that relying solely on textual data is insufficient. By effectively combining both textual and visual cues, LayoutXLM significantly improves the performance and comprehensiveness of document analysis tasks.

A second implication of our analysis is related to the deployability of advanced NLP models. One often faces a compromise between efficient model training and performance. More specifically, training or fine-tuning of a (NLP) model can be a major obstacle, as it requires large amounts of labeled data, which is rare and costly to acquire. In the context of this study, however, we observe the ability of LayoutXML to extract information from visually rich banking documents without requiring a lot of labeled training data and also without requiring a lot of fine-tuning. These features substantially reduce the effort associated with using LayoutXML in practice.

Relatedly, the problem of token class imbalance in banking documents is a major challenge. LayoutXLM handles this problem without the need for manual data balancing or preprocessing, which emphasizes the model's ability to adapt to the inherent complexity of banking documents. This ensures an authentic representation of document data during analysis and, again, reduces deployment times and resources.

Concerning the field of NLP, our study questions an exclusive use of zero-shot learning, even when using large language models such as GPT-3.5. Instead, these document analytics frameworks should currently be approached as few-shot learners. The amount of knowledge encapsulated in pre-trained models may not capture the intricate nuances of specialized domains such as banking.
As of now, minimal fine-tuning remains necessary to adapt models to the unique characteristics and challenges of banking documents to ensure optimal performance and accurate information extraction. In the future, this dynamic may change as NLP continues to evolve.\\

Our study also shows limitations that suggest avenues for future research. In particular, handling banking documents with extreme layout variability remains a challenge. Other multimodal models such as GPT-4 have attracted much attention and yet, their lack of awareness of layout is a shortcoming. Thus, a direct comparison between a layout-aware variant of GPT and LayoutXLM would be a fruitful route for future research.

 \bibliographystyle{elsarticle-harv} 
 \bibliography{literature}

\begin{thebibliography}{61}
\expandafter\ifx\csname natexlab\endcsname\relax\def\natexlab#1{#1}\fi
\providecommand{\url}[1]{\texttt{#1}}
\providecommand{\href}[2]{#2}
\providecommand{\path}[1]{#1}
\providecommand{\DOIprefix}{doi:}
\providecommand{\ArXivprefix}{arXiv:}
\providecommand{\URLprefix}{URL: }
\providecommand{\Pubmedprefix}{pmid:}
\providecommand{\doi}[1]{\href{http://dx.doi.org/#1}{\path{#1}}}
\providecommand{\Pubmed}[1]{\href{pmid:#1}{\path{#1}}}
\providecommand{\bibinfo}[2]{#2}
\ifx\xfnm\relax \def\xfnm[#1]{\unskip,\space#1}\fi
\bibitem[{Aureli(2017)}]{aureli2017}
\bibinfo{author}{Aureli, S.}, \bibinfo{year}{2017}.
\newblock \bibinfo{title}{A comparison of content analysis usage and text
  mining in csr corporate disclosure}.
\newblock \bibinfo{journal}{The International Journal of Digital Accounting
  Research} \bibinfo{volume}{17}, \bibinfo{pages}{1--32}.
\newblock \DOIprefix\doi{10.4192/1577-8517-v17_1}.
\bibitem[{Baviskar et~al.(2021)Baviskar, Ahirrao, Potdar and
  Kotecha}]{baviskar2021}
\bibinfo{author}{Baviskar, D.}, \bibinfo{author}{Ahirrao, S.},
  \bibinfo{author}{Potdar, V.}, \bibinfo{author}{Kotecha, K.},
  \bibinfo{year}{2021}.
\newblock \bibinfo{title}{Efficient automated processing of the unstructured
  documents using artificial intelligence: A systematic literature review and
  future directions}.
\newblock \bibinfo{journal}{IEEE Access} \bibinfo{volume}{9},
  \bibinfo{pages}{72894--72936}.
\newblock \DOIprefix\doi{10.1109/ACCESS.2021.3072900}.
\bibitem[{Boulahia et~al.(2021)Boulahia, Amamra, Madi and Daikh}]{boulahia2021}
\bibinfo{author}{Boulahia, S.}, \bibinfo{author}{Amamra, A.},
  \bibinfo{author}{Madi, M.}, \bibinfo{author}{Daikh, S.},
  \bibinfo{year}{2021}.
\newblock \bibinfo{title}{Early, intermediate and late fusion strategies for
  robust deep learning-based multimodal action recognition}.
\newblock \bibinfo{journal}{Machine Vision and Applications}
  \bibinfo{volume}{32}.
\newblock \DOIprefix\doi{10.1007/s00138-021-01249-8}.
\bibitem[{Brown et~al.(2020)Brown, Mann, Ryder, Subbiah, Kaplan, Dhariwal,
  Neelakantan, Shyam, Sastry, Askell, Agarwal, Herbert-Voss, Krueger, Henighan,
  Child, Ramesh, Ziegler, Wu, Winter, Hesse, Chen, Sigler, Litwin, Gray, Chess,
  Clark, Berner, McCandlish, Radford, Sutskever and Amodei}]{brown2020language}
\bibinfo{author}{Brown, T.B.}, \bibinfo{author}{Mann, B.},
  \bibinfo{author}{Ryder, N.}, \bibinfo{author}{Subbiah, M.},
  \bibinfo{author}{Kaplan, J.}, \bibinfo{author}{Dhariwal, P.},
  \bibinfo{author}{Neelakantan, A.}, \bibinfo{author}{Shyam, P.},
  \bibinfo{author}{Sastry, G.}, \bibinfo{author}{Askell, A.},
  \bibinfo{author}{Agarwal, S.}, \bibinfo{author}{Herbert-Voss, A.},
  \bibinfo{author}{Krueger, G.}, \bibinfo{author}{Henighan, T.},
  \bibinfo{author}{Child, R.}, \bibinfo{author}{Ramesh, A.},
  \bibinfo{author}{Ziegler, D.M.}, \bibinfo{author}{Wu, J.},
  \bibinfo{author}{Winter, C.}, \bibinfo{author}{Hesse, C.},
  \bibinfo{author}{Chen, M.}, \bibinfo{author}{Sigler, E.},
  \bibinfo{author}{Litwin, M.}, \bibinfo{author}{Gray, S.},
  \bibinfo{author}{Chess, B.}, \bibinfo{author}{Clark, J.},
  \bibinfo{author}{Berner, C.}, \bibinfo{author}{McCandlish, S.},
  \bibinfo{author}{Radford, A.}, \bibinfo{author}{Sutskever, I.},
  \bibinfo{author}{Amodei, D.}, \bibinfo{year}{2020}.
\newblock \bibinfo{title}{Language models are few-shot learners}.
\newblock \href{http://arxiv.org/abs/2005.14165}{{\tt arXiv:2005.14165}}.
\bibitem[{Casu et~al.(2015)Casu, Girardone and Molyneux}]{casu2015introduction}
\bibinfo{author}{Casu, B.}, \bibinfo{author}{Girardone, C.},
  \bibinfo{author}{Molyneux, P.}, \bibinfo{year}{2015}.
\newblock \bibinfo{title}{Introduction to Banking}.
\newblock \bibinfo{publisher}{Pearson}.
\bibitem[{Chaturvedi and Chopra(2014)}]{Chaturvedi2014}
\bibinfo{author}{Chaturvedi, D.}, \bibinfo{author}{Chopra, S.},
  \bibinfo{year}{2014}.
\newblock \bibinfo{title}{Customers sentiment on banks}.
\newblock \bibinfo{journal}{International Journal of Computer Applications}
  \bibinfo{volume}{98}, \bibinfo{pages}{8--13}.
\newblock \DOIprefix\doi{10.5120/17242-7578}.
\bibitem[{Chen et~al.(2019)Chen, Huang and Chen}]{chen2019}
\bibinfo{author}{Chen, C.C.}, \bibinfo{author}{Huang, H.H.},
  \bibinfo{author}{Chen, H.H.}, \bibinfo{year}{2019}.
\newblock \bibinfo{title}{Next cashtag prediction on social trading platforms
  with auxiliary tasks}, in: \bibinfo{booktitle}{2019 IEEE/ACM International
  Conference on Advances in Social Networks Analysis and Mining (ASONAM)}, pp.
  \bibinfo{pages}{525--527}.
\newblock \DOIprefix\doi{10.1145/3341161.3342945}.
\bibitem[{Chen et~al.(2020)Chen, Huang and Chen}]{chen2020}
\bibinfo{author}{Chen, C.C.}, \bibinfo{author}{Huang, H.H.},
  \bibinfo{author}{Chen, H.H.}, \bibinfo{year}{2020}.
\newblock \bibinfo{title}{Nlp in fintech applications: Past, present and
  future}.
\newblock \href{http://arxiv.org/abs/2005.01320}{{\tt arXiv:2005.01320}}.
\bibitem[{Chen et~al.(2022)Chen, Lv, Cui, Zhang and Wei}]{chenXDoc}
\bibinfo{author}{Chen, J.}, \bibinfo{author}{Lv, T.}, \bibinfo{author}{Cui,
  L.}, \bibinfo{author}{Zhang, C.}, \bibinfo{author}{Wei, F.},
  \bibinfo{year}{2022}.
\newblock \bibinfo{title}{{XD}oc: Unified pre-training for cross-format
  document understanding}, in: \bibinfo{booktitle}{Findings of the Association
  for Computational Linguistics: EMNLP 2022}, \bibinfo{publisher}{Association
  for Computational Linguistics}, \bibinfo{address}{Abu Dhabi, United Arab
  Emirates}. pp. \bibinfo{pages}{1006--1016}.
\newblock \URLprefix \url{https://aclanthology.org/2022.findings-emnlp.71}.
\bibitem[{Chi et~al.(2021)Chi, Dong, Wei, Yang, Singhal, Wang, Song, Mao, Huang
  and Zhou}]{chiInfoXLM}
\bibinfo{author}{Chi, Z.}, \bibinfo{author}{Dong, L.}, \bibinfo{author}{Wei,
  F.}, \bibinfo{author}{Yang, N.}, \bibinfo{author}{Singhal, S.},
  \bibinfo{author}{Wang, W.}, \bibinfo{author}{Song, X.}, \bibinfo{author}{Mao,
  X.L.}, \bibinfo{author}{Huang, H.}, \bibinfo{author}{Zhou, M.},
  \bibinfo{year}{2021}.
\newblock \bibinfo{title}{{I}nfo{XLM}: An information-theoretic framework for
  cross-lingual language model pre-training}, in:
  \bibinfo{booktitle}{Proceedings of the 2021 Conference of the North American
  Chapter of the Association for Computational Linguistics: Human Language
  Technologies}, \bibinfo{publisher}{Association for Computational
  Linguistics}, \bibinfo{address}{Online}. pp. \bibinfo{pages}{3576--3588}.
\newblock \DOIprefix\doi{10.18653/v1/2021.naacl-main.280}.
\bibitem[{Conneau et~al.(2020)Conneau, Khandelwal, Goyal, Chaudhary, Wenzek,
  Guzm{\'a}n, Grave, Ott, Zettlemoyer and Stoyanov}]{conneauXLMRoberta}
\bibinfo{author}{Conneau, A.}, \bibinfo{author}{Khandelwal, K.},
  \bibinfo{author}{Goyal, N.}, \bibinfo{author}{Chaudhary, V.},
  \bibinfo{author}{Wenzek, G.}, \bibinfo{author}{Guzm{\'a}n, F.},
  \bibinfo{author}{Grave, E.}, \bibinfo{author}{Ott, M.},
  \bibinfo{author}{Zettlemoyer, L.}, \bibinfo{author}{Stoyanov, V.},
  \bibinfo{year}{2020}.
\newblock \bibinfo{title}{Unsupervised cross-lingual representation learning at
  scale}, in: \bibinfo{booktitle}{Proceedings of the 58th Annual Meeting of the
  Association for Computational Linguistics}, \bibinfo{publisher}{Association
  for Computational Linguistics}, \bibinfo{address}{Online}. pp.
  \bibinfo{pages}{8440--8451}.
\newblock \DOIprefix\doi{10.18653/v1/2020.acl-main.747}.
\bibitem[{Conneau and Lample(2019)}]{conneauXLM}
\bibinfo{author}{Conneau, A.}, \bibinfo{author}{Lample, G.},
  \bibinfo{year}{2019}.
\newblock \bibinfo{title}{Cross-lingual language model pretraining}, in:
  \bibinfo{editor}{Wallach, H.}, \bibinfo{editor}{Larochelle, H.},
  \bibinfo{editor}{Beygelzimer, A.}, \bibinfo{editor}{d\textquotesingle
  Alch\'{e}-Buc, F.}, \bibinfo{editor}{Fox, E.}, \bibinfo{editor}{Garnett, R.}
  (Eds.), \bibinfo{booktitle}{Advances in Neural Information Processing
  Systems}, \bibinfo{publisher}{Curran Associates, Inc.}. pp.
  \bibinfo{pages}{7059--7069}.
\newblock \URLprefix
  \url{https://proceedings.neurips.cc/paper_files/paper/2019/file/c04c19c2c2474dbf5f7ac4372c5b9af1-Paper.pdf}.
\bibitem[{Craja et~al.(2020)Craja, Kim and Lessmann}]{craja2020}
\bibinfo{author}{Craja, P.}, \bibinfo{author}{Kim, A.},
  \bibinfo{author}{Lessmann, S.}, \bibinfo{year}{2020}.
\newblock \bibinfo{title}{Deep learning for detecting financial statement
  fraud}.
\newblock \bibinfo{journal}{Decision Support Systems} \bibinfo{volume}{139},
  \bibinfo{pages}{113421}.
\newblock \DOIprefix\doi{10.1016/j.dss.2020.113421}.
\bibitem[{Dang et~al.(2018)Dang, Sadeghi-Niaraki, Huynh, Min and
  Moon}]{dang2018}
\bibinfo{author}{Dang, L.M.}, \bibinfo{author}{Sadeghi-Niaraki, A.},
  \bibinfo{author}{Huynh, H.}, \bibinfo{author}{Min, K.},
  \bibinfo{author}{Moon, H.}, \bibinfo{year}{2018}.
\newblock \bibinfo{title}{Deep learning approach for short-term stock trends
  prediction based on two-stream gated recurrent unit network}.
\newblock \bibinfo{journal}{IEEE Access} \bibinfo{volume}{PP},
  \bibinfo{pages}{1--1}.
\newblock \DOIprefix\doi{10.1109/ACCESS.2018.2868970}.
\bibitem[{{De Caigny} et~al.(2020){De Caigny}, Coussement, {De Bock} and
  Lessmann}]{decaigny2020}
\bibinfo{author}{{De Caigny}, A.}, \bibinfo{author}{Coussement, K.},
  \bibinfo{author}{{De Bock}, K.W.}, \bibinfo{author}{Lessmann, S.},
  \bibinfo{year}{2020}.
\newblock \bibinfo{title}{Incorporating textual information in customer churn
  prediction models based on a convolutional neural network}.
\newblock \bibinfo{journal}{International Journal of Forecasting}
  \bibinfo{volume}{36}, \bibinfo{pages}{1563--1578}.
\newblock \DOIprefix\doi{10.1016/j.ijforecast.2019.03.029}.
\bibitem[{Devlin et~al.(2018)Devlin, Chang, Lee and Toutanova}]{devlinBert}
\bibinfo{author}{Devlin, J.}, \bibinfo{author}{Chang, M.},
  \bibinfo{author}{Lee, K.}, \bibinfo{author}{Toutanova, K.},
  \bibinfo{year}{2018}.
\newblock \bibinfo{title}{{BERT:} pre-training of deep bidirectional
  transformers for language understanding}.
\newblock \bibinfo{journal}{CoRR} \bibinfo{volume}{abs/1810.04805}.
\newblock \href{http://arxiv.org/abs/1810.04805}{{\tt arXiv:1810.04805}}.
\bibitem[{Dong(2023)}]{dong2023}
\bibinfo{author}{Dong, Y.}, \bibinfo{year}{2023}.
\newblock \bibinfo{title}{Esg scoring system construction: Portfolio investment
  based on machine learning}.
\newblock \bibinfo{journal}{Advances in Economics, Management and Political
  Sciences} \bibinfo{volume}{3}, \bibinfo{pages}{517--525}.
\newblock \DOIprefix\doi{10.54254/2754-1169/3/2022829}.
\bibitem[{Doumpos et~al.(2023)Doumpos, Zopounidis, Gounopoulos, Platanakis and
  Zhang}]{doumpos2023}
\bibinfo{author}{Doumpos, M.}, \bibinfo{author}{Zopounidis, C.},
  \bibinfo{author}{Gounopoulos, D.}, \bibinfo{author}{Platanakis, E.},
  \bibinfo{author}{Zhang, W.}, \bibinfo{year}{2023}.
\newblock \bibinfo{title}{Operational research and artificial intelligence
  methods in banking}.
\newblock \bibinfo{journal}{European Journal of Operational Research}
  \bibinfo{volume}{306}, \bibinfo{pages}{1--16}.
\newblock \DOIprefix\doi{10.1016/j.ejor.2022.04.027}.
\bibitem[{Engin et~al.(2019)Engin, Emekligil, Akpinar, Oral and
  Arslan}]{engin2019}
\bibinfo{author}{Engin, D.}, \bibinfo{author}{Emekligil, E.},
  \bibinfo{author}{Akpinar, M.Y.}, \bibinfo{author}{Oral, B.},
  \bibinfo{author}{Arslan, S.}, \bibinfo{year}{2019}.
\newblock \bibinfo{title}{Multimodal deep neural networks for banking document
  classiﬁcation}.
\bibitem[{Fernández~Rodríguez et~al.(2022)Fernández~Rodríguez, Papale,
  Carminati and Zanero}]{rodriguez2022}
\bibinfo{author}{Fernández~Rodríguez, J.}, \bibinfo{author}{Papale, M.},
  \bibinfo{author}{Carminati, M.}, \bibinfo{author}{Zanero, S.},
  \bibinfo{year}{2022}.
\newblock \bibinfo{title}{A natural language processing approach for financial
  fraud detection}, in: \bibinfo{editor}{Demetrescu, C.}, \bibinfo{editor}{Mei,
  A.} (Eds.), \bibinfo{booktitle}{Proceedings of the Italian Conference on
  Cybersecurity ITASEC 2022}, \bibinfo{publisher}{CEUR-WS.org},
  \bibinfo{address}{Rome, Italy}. pp. \bibinfo{pages}{135--149}.
\newblock \URLprefix \url{http://ceur-ws.org/Vol-3260/paper10.pdf}.
\bibitem[{Fisher et~al.(2016)Fisher, Garnsey and Hughes}]{fisher2016}
\bibinfo{author}{Fisher, I.}, \bibinfo{author}{Garnsey, M.},
  \bibinfo{author}{Hughes, M.}, \bibinfo{year}{2016}.
\newblock \bibinfo{title}{Natural language processing in accounting, auditing
  and finance: A synthesis of the literature with a roadmap for future
  research}.
\newblock \bibinfo{journal}{Intelligent Systems in Accounting, Finance and
  Management} \bibinfo{volume}{23}, \bibinfo{pages}{157–214}.
\newblock \DOIprefix\doi{10.1002/isaf.1386}.
\bibitem[{Gerling(2023)}]{gerling2023}
\bibinfo{author}{Gerling, C.}, \bibinfo{year}{2023}.
\newblock \bibinfo{title}{Company2{V}ec - {G}erman company embeddings based on
  corporate websites}.
\newblock \bibinfo{journal}{International Journal of Information Technology \&
  Decision Making} \DOIprefix\doi{10.1142/S0219622023500694}.
\bibitem[{Gupta et~al.(2020)Gupta, Dengre, Kheruwala and Shah}]{gupta2020}
\bibinfo{author}{Gupta, A.}, \bibinfo{author}{Dengre, V.},
  \bibinfo{author}{Kheruwala, H.}, \bibinfo{author}{Shah, M.},
  \bibinfo{year}{2020}.
\newblock \bibinfo{title}{Comprehensive review of text-mining applications in
  finance}.
\newblock \bibinfo{journal}{Journal of Financial Innovation}
  \bibinfo{volume}{6}.
\newblock \DOIprefix\doi{10.1186/s40854-020-00205-1}.
\bibitem[{Huang et~al.(2022)Huang, Lv, Cui, Lu and Wei}]{huang2022layoutlmv3}
\bibinfo{author}{Huang, Y.}, \bibinfo{author}{Lv, T.}, \bibinfo{author}{Cui,
  L.}, \bibinfo{author}{Lu, Y.}, \bibinfo{author}{Wei, F.},
  \bibinfo{year}{2022}.
\newblock \bibinfo{title}{Layout{LM}v3: Pre-training for document ai with
  unified text and image masking}.
\newblock \href{http://arxiv.org/abs/2204.08387}{{\tt arXiv:2204.08387}}.
\bibitem[{Jiang et~al.(2018)Jiang, Wang, Wang and Ding}]{jiang2018}
\bibinfo{author}{Jiang, C.}, \bibinfo{author}{Wang, Z.}, \bibinfo{author}{Wang,
  R.}, \bibinfo{author}{Ding, Y.}, \bibinfo{year}{2018}.
\newblock \bibinfo{title}{{Loan default prediction by combining soft
  information extracted from descriptive text in online peer-to-peer lending}}.
\newblock \bibinfo{journal}{Annals of Operations Research}
  \bibinfo{volume}{266}, \bibinfo{pages}{511--529}.
\newblock \DOIprefix\doi{10.1007/s10479-017-2668-z}.
\bibitem[{Kamaruddin et~al.(2015)Kamaruddin, Bakar, Hamdan, Nor, Nazri, Othman
  and Hussein}]{kamaruddin2015}
\bibinfo{author}{Kamaruddin, S.S.}, \bibinfo{author}{Bakar, A.A.},
  \bibinfo{author}{Hamdan, A.R.}, \bibinfo{author}{Nor, F.M.},
  \bibinfo{author}{Nazri, M.Z.A.}, \bibinfo{author}{Othman, Z.A.},
  \bibinfo{author}{Hussein, G.S.}, \bibinfo{year}{2015}.
\newblock \bibinfo{title}{A text mining system for deviation detection in
  financial documents}.
\newblock \bibinfo{journal}{Intelligent Data Analysis} \bibinfo{volume}{19},
  \bibinfo{pages}{19--44}.
\newblock \DOIprefix\doi{10.3233/IDA-150768}.
\bibitem[{Kotios et~al.(2022)Kotios, Makridis, Fatouros and
  Kyriazis}]{kotios2022}
\bibinfo{author}{Kotios, D.}, \bibinfo{author}{Makridis, G.},
  \bibinfo{author}{Fatouros, G.}, \bibinfo{author}{Kyriazis, D.},
  \bibinfo{year}{2022}.
\newblock \bibinfo{title}{Deep learning enhancing banking services: a hybrid
  transaction classification and cash flow prediction approach}.
\newblock \bibinfo{journal}{Journal of Big Data} \bibinfo{volume}{9},
  \bibinfo{pages}{100}.
\newblock \DOIprefix\doi{10.1186/s40537-022-00651-x}.
\bibitem[{Kumar and Ravi(2016)}]{kumar2016}
\bibinfo{author}{Kumar, B.S.}, \bibinfo{author}{Ravi, V.},
  \bibinfo{year}{2016}.
\newblock \bibinfo{title}{A survey of the applications of text mining in
  financial domain}.
\newblock \bibinfo{journal}{Knowledge-Based Systems} \bibinfo{volume}{114},
  \bibinfo{pages}{128--147}.
\newblock \DOIprefix\doi{10.1016/j.knosys.2016.10.003}.
\bibitem[{Lee and Yoo(2019)}]{lee2019}
\bibinfo{author}{Lee, S.I.}, \bibinfo{author}{Yoo, S.J.}, \bibinfo{year}{2019}.
\newblock \bibinfo{title}{Multimodal deep learning for finance: Integrating and
  forecasting international stock markets}.
\newblock \href{http://arxiv.org/abs/1903.06478}{{\tt arXiv:1903.06478}}.
\bibitem[{Levich et~al.(2023)Levich, Lutz and Neumann}]{levich2023}
\bibinfo{author}{Levich, S.}, \bibinfo{author}{Lutz, B.},
  \bibinfo{author}{Neumann, D.}, \bibinfo{year}{2023}.
\newblock \bibinfo{title}{Utilizing the omnipresent: Incorporating digital
  documents into predictive process monitoring using deep neural networks}.
\newblock \bibinfo{journal}{Decision Support Systems} ,
  \bibinfo{pages}{114043}\DOIprefix\doi{10.1016/j.dss.2023.114043}.
\bibitem[{Lewis and Young(2019)}]{lewis2019}
\bibinfo{author}{Lewis, C.M.}, \bibinfo{author}{Young, S.},
  \bibinfo{year}{2019}.
\newblock \bibinfo{title}{Fad or future? automated analysis of financial text
  and its implications for corporate reporting}.
\newblock \bibinfo{journal}{Accounting and Business Research}
  \bibinfo{volume}{49}, \bibinfo{pages}{587 -- 615}.
\newblock \DOIprefix\doi{10.1080/00014788.2019.1611730}.
\bibitem[{Li(2011)}]{li2011}
\bibinfo{author}{Li, F.}, \bibinfo{year}{2011}.
\newblock \bibinfo{title}{Textual analysis of corporate disclosures: A survey
  of the literature}.
\newblock \bibinfo{journal}{Journal of Accounting Literature}
  \bibinfo{volume}{29}, \bibinfo{pages}{143--167}.
\bibitem[{Li et~al.(2022a)Li, Xu, Cui and Wei}]{liMarkupLM}
\bibinfo{author}{Li, J.}, \bibinfo{author}{Xu, Y.}, \bibinfo{author}{Cui, L.},
  \bibinfo{author}{Wei, F.}, \bibinfo{year}{2022}a.
\newblock \bibinfo{title}{{M}arkup{LM}: Pre-training of text and markup
  language for visually rich document understanding}, in:
  \bibinfo{booktitle}{Proceedings of the 60th Annual Meeting of the Association
  for Computational Linguistics (Volume 1: Long Papers)},
  \bibinfo{publisher}{Association for Computational Linguistics},
  \bibinfo{address}{Dublin, Ireland}. pp. \bibinfo{pages}{6078--6087}.
\newblock \DOIprefix\doi{10.18653/v1/2022.acl-long.420}.
\bibitem[{Li et~al.(2022b)Li, Xu, Lv, Cui, Zhang and Wei}]{liDIT}
\bibinfo{author}{Li, J.}, \bibinfo{author}{Xu, Y.}, \bibinfo{author}{Lv, T.},
  \bibinfo{author}{Cui, L.}, \bibinfo{author}{Zhang, C.}, \bibinfo{author}{Wei,
  F.}, \bibinfo{year}{2022}b.
\newblock \bibinfo{title}{Dit: Self-supervised pre-training for document image
  transformer}.
\newblock \href{http://arxiv.org/abs/2203.02378}{{\tt arXiv:2203.02378}}.
\bibitem[{Li et~al.(2021a)Li, Gu, Kuen, Morariu, Zhao, Jain, Manjunatha and
  Liu}]{liSelfDoc}
\bibinfo{author}{Li, P.}, \bibinfo{author}{Gu, J.}, \bibinfo{author}{Kuen, J.},
  \bibinfo{author}{Morariu, V.I.}, \bibinfo{author}{Zhao, H.},
  \bibinfo{author}{Jain, R.}, \bibinfo{author}{Manjunatha, V.},
  \bibinfo{author}{Liu, H.}, \bibinfo{year}{2021}a.
\newblock \bibinfo{title}{Selfdoc: Self-supervised document representation
  learning}, in: \bibinfo{booktitle}{2021 IEEE/CVF Conference on Computer
  Vision and Pattern Recognition (CVPR)}, pp. \bibinfo{pages}{5648--5656}.
\newblock \DOIprefix\doi{10.1109/CVPR46437.2021.00560}.
\bibitem[{Li et~al.(2021b)Li, Tan, Wang and Chen}]{li2021}
\bibinfo{author}{Li, Q.}, \bibinfo{author}{Tan, J.}, \bibinfo{author}{Wang,
  J.}, \bibinfo{author}{Chen, H.}, \bibinfo{year}{2021}b.
\newblock \bibinfo{title}{A multimodal event-driven lstm model for stock
  prediction using online news}.
\newblock \bibinfo{journal}{IEEE Transactions on Knowledge and Data
  Engineering} \bibinfo{volume}{33}, \bibinfo{pages}{3323--3337}.
\newblock \DOIprefix\doi{10.1109/TKDE.2020.2968894}.
\bibitem[{Mai et~al.(2019)Mai, Tian, Lee and Ma}]{mai2019}
\bibinfo{author}{Mai, F.}, \bibinfo{author}{Tian, S.}, \bibinfo{author}{Lee,
  C.}, \bibinfo{author}{Ma, L.}, \bibinfo{year}{2019}.
\newblock \bibinfo{title}{Deep learning models for bankruptcy prediction using
  textual disclosures}.
\newblock \bibinfo{journal}{European Journal of Operational Research}
  \bibinfo{volume}{274}, \bibinfo{pages}{743--758}.
\newblock \DOIprefix\doi{10.1016/j.ejor.2018.10.024}.
\bibitem[{Mikolov et~al.(2013)Mikolov, Chen, Corrado and
  Dean}]{mikolov2013efficient}
\bibinfo{author}{Mikolov, T.}, \bibinfo{author}{Chen, K.},
  \bibinfo{author}{Corrado, G.}, \bibinfo{author}{Dean, J.},
  \bibinfo{year}{2013}.
\newblock \bibinfo{title}{Efficient estimation of word representations in
  vector space}.
\newblock \href{http://arxiv.org/abs/1301.3781}{{\tt arXiv:1301.3781}}.
\bibitem[{Mogaji et~al.(2021)Mogaji, Balakrishnan, Nwoba and
  Nguyen}]{mogaji2021}
\bibinfo{author}{Mogaji, E.}, \bibinfo{author}{Balakrishnan, J.},
  \bibinfo{author}{Nwoba, A.C.}, \bibinfo{author}{Nguyen, N.P.},
  \bibinfo{year}{2021}.
\newblock \bibinfo{title}{Emerging-market consumers’ interactions with
  banking chatbots}.
\newblock \bibinfo{journal}{Telematics and Informatics} \bibinfo{volume}{65},
  \bibinfo{pages}{101711}.
\newblock \DOIprefix\doi{10.1016/j.tele.2021.101711}.
\bibitem[{Netzer et~al.(2019)Netzer, Lemaire and Herzenstein}]{netzer2019}
\bibinfo{author}{Netzer, O.}, \bibinfo{author}{Lemaire, A.},
  \bibinfo{author}{Herzenstein, M.}, \bibinfo{year}{2019}.
\newblock \bibinfo{title}{When words sweat: Identifying signals for loan
  default in the text of loan applications}.
\newblock \bibinfo{journal}{Journal of Marketing Research}
  \bibinfo{volume}{56}, \bibinfo{pages}{960--980}.
\newblock \DOIprefix\doi{10.1177/0022243719852959}.
\bibitem[{OpenAI(2023)}]{openai2023gpt4}
\bibinfo{author}{OpenAI}, \bibinfo{year}{2023}.
\newblock \bibinfo{title}{Gpt-4 technical report}.
\newblock \href{http://arxiv.org/abs/2303.08774}{{\tt arXiv:2303.08774}}.
\bibitem[{Oral et~al.(2019)Oral, Emekligil, Arslan and
  Eryi{\u{g}}it}]{oral2019}
\bibinfo{author}{Oral, B.}, \bibinfo{author}{Emekligil, E.},
  \bibinfo{author}{Arslan, S.}, \bibinfo{author}{Eryi{\u{g}}it, G.},
  \bibinfo{year}{2019}.
\newblock \bibinfo{title}{Extracting complex relations from banking documents},
  in: \bibinfo{booktitle}{Proceedings of the Second Workshop on Economics and
  Natural Language Processing}, \bibinfo{publisher}{Association for
  Computational Linguistics}, \bibinfo{address}{Hong Kong}. pp.
  \bibinfo{pages}{1--9}.
\newblock \DOIprefix\doi{10.18653/v1/D19-5101}.
\bibitem[{Oral et~al.(2020)Oral, Emekligil, Arslan and Eryiǧit}]{oral2020}
\bibinfo{author}{Oral, B.}, \bibinfo{author}{Emekligil, E.},
  \bibinfo{author}{Arslan, S.}, \bibinfo{author}{Eryiǧit, G.},
  \bibinfo{year}{2020}.
\newblock \bibinfo{title}{Information extraction from text intensive and
  visually rich banking documents}.
\newblock \bibinfo{journal}{Information Processing \& Management}
  \bibinfo{volume}{57}, \bibinfo{pages}{102361}.
\newblock \DOIprefix\doi{10.1016/j.ipm.2020.102361}.
\bibitem[{Oral and Eryi\u{g}it(2022)}]{oral2022}
\bibinfo{author}{Oral, B.}, \bibinfo{author}{Eryi\u{g}it, G.},
  \bibinfo{year}{2022}.
\newblock \bibinfo{title}{Fusion of visual representations for multimodal
  information extraction from unstructured transactional documents}.
\newblock \bibinfo{journal}{Int. J. Doc. Anal. Recognit.} \bibinfo{volume}{25},
  \bibinfo{pages}{187–205}.
\newblock \DOIprefix\doi{10.1007/s10032-022-00399-3}.
\bibitem[{Pejic~Bach et~al.(2019)Pejic~Bach, Krstic, Seljan and
  Turulja}]{pejic2019}
\bibinfo{author}{Pejic~Bach, M.}, \bibinfo{author}{Krstic, Z.},
  \bibinfo{author}{Seljan, S.}, \bibinfo{author}{Turulja, L.},
  \bibinfo{year}{2019}.
\newblock \bibinfo{title}{Text mining for big data analysis in financial
  sector: A literature review}.
\newblock \bibinfo{journal}{Sustainability} \bibinfo{volume}{11},
  \bibinfo{pages}{1277}.
\newblock \DOIprefix\doi{10.3390/su11051277}.
\bibitem[{Petersson et~al.(2023)Petersson, Pawar and
  Fagerstrøm}]{petersson2023}
\bibinfo{author}{Petersson, A.H.}, \bibinfo{author}{Pawar, S.},
  \bibinfo{author}{Fagerstrøm, A.}, \bibinfo{year}{2023}.
\newblock \bibinfo{title}{Investigating the factors of customer experiences
  using real-life text-based banking chatbot: a qualitative study in norway}.
\newblock \bibinfo{journal}{Procedia Computer Science} \bibinfo{volume}{219},
  \bibinfo{pages}{697--704}.
\newblock \DOIprefix\doi{10.1016/j.procs.2023.01.341}.
\bibitem[{Poria et~al.(2017)Poria, Cambria, Bajpai and Hussain}]{poria2017}
\bibinfo{author}{Poria, S.}, \bibinfo{author}{Cambria, E.},
  \bibinfo{author}{Bajpai, R.}, \bibinfo{author}{Hussain, A.},
  \bibinfo{year}{2017}.
\newblock \bibinfo{title}{A review of affective computing: From unimodal
  analysis to multimodal fusion}.
\newblock \bibinfo{journal}{Information Fusion} \bibinfo{volume}{37},
  \bibinfo{pages}{98--125}.
\newblock \DOIprefix\doi{10.1016/j.inffus.2017.02.003}.
\bibitem[{Schmitz et~al.(2023)Schmitz, Lutz, Wolff and Neumann}]{schmitz2023}
\bibinfo{author}{Schmitz, H.C.}, \bibinfo{author}{Lutz, B.},
  \bibinfo{author}{Wolff, D.}, \bibinfo{author}{Neumann, D.},
  \bibinfo{year}{2023}.
\newblock \bibinfo{title}{When machines trade on corporate disclosures: Using
  text analytics for investment strategies}.
\newblock \bibinfo{journal}{Decision Support Systems} \bibinfo{volume}{165},
  \bibinfo{pages}{113892}.
\newblock \DOIprefix\doi{https://doi.org/10.1016/j.dss.2022.113892}.
\bibitem[{Sokolov et~al.(2021)Sokolov, Mostovoy, Ding and Seco}]{sokolov2021}
\bibinfo{author}{Sokolov, A.}, \bibinfo{author}{Mostovoy, J.},
  \bibinfo{author}{Ding, J.}, \bibinfo{author}{Seco, L.}, \bibinfo{year}{2021}.
\newblock \bibinfo{title}{Building machine learning systems for automated esg
  scoring}.
\newblock \bibinfo{journal}{The Journal of Impact and ESG Investing}
  \bibinfo{volume}{1}, \bibinfo{pages}{39--50}.
\newblock \DOIprefix\doi{10.3905/jesg.2021.1.010}.
\bibitem[{Stevenson et~al.(2021)Stevenson, Mues and Bravo}]{stevenson2021}
\bibinfo{author}{Stevenson, M.}, \bibinfo{author}{Mues, C.},
  \bibinfo{author}{Bravo, C.}, \bibinfo{year}{2021}.
\newblock \bibinfo{title}{The value of text for small business default
  prediction: A deep learning approach}.
\newblock \bibinfo{journal}{European Journal of Operational Research}
  \bibinfo{volume}{295}, \bibinfo{pages}{758--771}.
\newblock \DOIprefix\doi{10.1016/j.ejor.2021.03.008}.
\bibitem[{Suhel et~al.(2020)Suhel, Shukla, Vyas and Mishra}]{suhel2020}
\bibinfo{author}{Suhel, S.F.}, \bibinfo{author}{Shukla, V.K.},
  \bibinfo{author}{Vyas, S.}, \bibinfo{author}{Mishra, V.P.},
  \bibinfo{year}{2020}.
\newblock \bibinfo{title}{Conversation to automation in banking through chatbot
  using artificial machine intelligence language}, in: \bibinfo{booktitle}{2020
  8th International Conference on Reliability, Infocom Technologies and
  Optimization (Trends and Future Directions) (ICRITO)}, pp.
  \bibinfo{pages}{611--618}.
\newblock \DOIprefix\doi{10.1109/ICRITO48877.2020.9197825}.
\bibitem[{Sumathi and Sheela(2017)}]{sumathi2017}
\bibinfo{author}{Sumathi, N.}, \bibinfo{author}{Sheela, T.},
  \bibinfo{year}{2017}.
\newblock \bibinfo{title}{Opinion mining analysis in banking system using rough
  feature selection technique from social media text}.
\newblock \bibinfo{journal}{International Journal of Mechanical Engineering and
  Technology} \bibinfo{volume}{8}, \bibinfo{pages}{274--289}.
\bibitem[{Sun et~al.(2018)Sun, Fang and Wang}]{sun2018}
\bibinfo{author}{Sun, Y.}, \bibinfo{author}{Fang, M.}, \bibinfo{author}{Wang,
  X.}, \bibinfo{year}{2018}.
\newblock \bibinfo{title}{A novel stock recommendation system using guba
  sentiment analysis}.
\newblock \bibinfo{journal}{Personal and Ubiquitous Computing}
  \bibinfo{volume}{22}.
\newblock \DOIprefix\doi{10.1007/s00779-018-1121-x}.
\bibitem[{Tavakoli et~al.(2023)Tavakoli, Chandra, Tian and
  Bravo}]{tavakoli2023}
\bibinfo{author}{Tavakoli, M.}, \bibinfo{author}{Chandra, R.},
  \bibinfo{author}{Tian, F.}, \bibinfo{author}{Bravo, C.},
  \bibinfo{year}{2023}.
\newblock \bibinfo{title}{Multi-modal deep learning for credit rating
  prediction using text and numerical data streams}.
\newblock \href{http://arxiv.org/abs/2304.10740}{{\tt arXiv:2304.10740}}.
\bibitem[{Tecles and Tabak(2010)}]{tecles2010}
\bibinfo{author}{Tecles, P.L.}, \bibinfo{author}{Tabak, B.M.},
  \bibinfo{year}{2010}.
\newblock \bibinfo{title}{Determinants of bank efficiency: The case of brazil}.
\newblock \bibinfo{journal}{European Journal of Operational Research}
  \bibinfo{volume}{207}, \bibinfo{pages}{1587--1598}.
\newblock \DOIprefix\doi{10.1016/j.ejor.2010.06.007}.
\bibitem[{Wang et~al.(2023)Wang, Ma and Chen}]{wang2023}
\bibinfo{author}{Wang, G.}, \bibinfo{author}{Ma, J.}, \bibinfo{author}{Chen,
  G.}, \bibinfo{year}{2023}.
\newblock \bibinfo{title}{Attentive statement fraud detection: Distinguishing
  multimodal financial data with fine-grained attention}.
\newblock \bibinfo{journal}{Decision Support Systems} \bibinfo{volume}{167},
  \bibinfo{pages}{113913}.
\newblock \DOIprefix\doi{https://doi.org/10.1016/j.dss.2022.113913}.
\bibitem[{Xing et~al.(2018)Xing, Cambria and Welsch}]{xing2018}
\bibinfo{author}{Xing, F.Z.}, \bibinfo{author}{Cambria, E.},
  \bibinfo{author}{Welsch, R.E.}, \bibinfo{year}{2018}.
\newblock \bibinfo{title}{Natural language based financial forecasting: A
  survey}.
\newblock \bibinfo{journal}{Artif. Intell. Rev.} \bibinfo{volume}{50},
  \bibinfo{pages}{49–73}.
\newblock \DOIprefix\doi{10.1007/s10462-017-9588-9}.
\bibitem[{Xu et~al.(2020)Xu, Li, Cui, Huang, Wei and Zhou}]{xuLayoutLMv1}
\bibinfo{author}{Xu, Y.}, \bibinfo{author}{Li, M.}, \bibinfo{author}{Cui, L.},
  \bibinfo{author}{Huang, S.}, \bibinfo{author}{Wei, F.},
  \bibinfo{author}{Zhou, M.}, \bibinfo{year}{2020}.
\newblock \bibinfo{title}{{LayoutLM}: Pre-training of text and layout for
  document image understanding}, in: \bibinfo{booktitle}{Proceedings of the
  26th {ACM} {SIGKDD} International Conference on Knowledge Discovery and Data
  Mining}, \bibinfo{publisher}{{ACM}}.
\newblock \DOIprefix\doi{10.1145/3394486.3403172}.
\bibitem[{Xu et~al.(2021a)Xu, Lv, Cui, Wang, Lu, Florencio, Zhang and
  Wei}]{xu2021layoutxlm}
\bibinfo{author}{Xu, Y.}, \bibinfo{author}{Lv, T.}, \bibinfo{author}{Cui, L.},
  \bibinfo{author}{Wang, G.}, \bibinfo{author}{Lu, Y.},
  \bibinfo{author}{Florencio, D.}, \bibinfo{author}{Zhang, C.},
  \bibinfo{author}{Wei, F.}, \bibinfo{year}{2021}a.
\newblock \bibinfo{title}{Layoutxlm: Multimodal pre-training for multilingual
  visually-rich document understanding}.
\newblock \href{http://arxiv.org/abs/2104.08836}{{\tt arXiv:2104.08836}}.
\bibitem[{Xu et~al.(2021b)Xu, Xu, Lv, Cui, Wei, Wang, Lu, Florencio, Zhang,
  Che, Zhang and Zhou}]{xuLayoutlmv2}
\bibinfo{author}{Xu, Y.}, \bibinfo{author}{Xu, Y.}, \bibinfo{author}{Lv, T.},
  \bibinfo{author}{Cui, L.}, \bibinfo{author}{Wei, F.}, \bibinfo{author}{Wang,
  G.}, \bibinfo{author}{Lu, Y.}, \bibinfo{author}{Florencio, D.},
  \bibinfo{author}{Zhang, C.}, \bibinfo{author}{Che, W.},
  \bibinfo{author}{Zhang, M.}, \bibinfo{author}{Zhou, L.},
  \bibinfo{year}{2021}b.
\newblock \bibinfo{title}{{L}ayout{LM}v2: Multi-modal pre-training for
  visually-rich document understanding}, in: \bibinfo{booktitle}{Proceedings of
  the 59th Annual Meeting of the Association for Computational Linguistics and
  the 11th International Joint Conference on Natural Language Processing
  (Volume 1: Long Papers)}, \bibinfo{publisher}{Association for Computational
  Linguistics}, \bibinfo{address}{Online}. pp. \bibinfo{pages}{2579--2591}.
\newblock \DOIprefix\doi{10.18653/v1/2021.acl-long.201}.
\bibitem[{Zhu et~al.(2015)Zhu, Chen and Guo}]{zhu2015}
\bibinfo{author}{Zhu, Y.}, \bibinfo{author}{Chen, W.}, \bibinfo{author}{Guo,
  G.}, \bibinfo{year}{2015}.
\newblock \bibinfo{title}{Fusing multiple features for depth-based action
  recognition}.
\newblock \bibinfo{journal}{ACM Trans. Intell. Syst. Technol.}
  \bibinfo{volume}{6}.
\newblock \DOIprefix\doi{10.1145/2629483}.

\end{thebibliography}

\end{document}